\documentclass[final]{cvpr}

\usepackage{times}
\usepackage{epsfig}
\usepackage{graphicx}
\usepackage{amsmath}
\usepackage{amssymb}
\usepackage{amsthm}
\usepackage{algorithm}
\usepackage{algorithmic}
\usepackage{siunitx}
\usepackage{textcomp}
\usepackage{pifont}% http://ctan.org/pkg/pifont
\newcommand{\cmark}{\ding{51}}%
\newcommand{\xmark}{\ding{55}}%
\usepackage{pifont}

\usepackage{dsfont}
\usepackage{array}
\usepackage{booktabs} 
\usepackage{listings}
\usepackage{xcolor}
\DeclareMathOperator{\E}{\mathbb{E}}
\newcommand{\norm}[1]{\left\lVert#1\right\rVert}
\newcommand{\bs}[1]{\boldsymbol{#1}}

\newtheorem{theorem}{Theorem}
\newtheorem{remark}{Remark}

\allowdisplaybreaks
\definecolor{codegreen}{rgb}{0,0.6,0}
\definecolor{codegray}{rgb}{0.5,0.5,0.5}
\definecolor{codepurple}{rgb}{0.58,0,0.82}
\definecolor{backcolour}{rgb}{0.95,0.95,0.92}

\lstdefinestyle{mystyle}{
    backgroundcolor=\color{backcolour},   
    commentstyle=\color{codegreen},
    keywordstyle=\color{magenta},
    numberstyle=\tiny\color{codegray},
    stringstyle=\color{codepurple},
    basicstyle=\ttfamily\footnotesize,
    breakatwhitespace=false,         
    breaklines=true,                 
    captionpos=b,                    
    keepspaces=true,                 
    numbers=left,                    
    numbersep=5pt,                  
    showspaces=false,                
    showstringspaces=false,
    showtabs=false,                  
    tabsize=2
}

\usepackage[pagebackref=true,breaklinks=true,colorlinks,bookmarks=false]{hyperref}

\def\cvprPaperID{11586} % *** Enter the CVPR Paper ID here
\def\confYear{CVPR 2021}

\begin{document}

%%%%%%%%% TITLE
\title{Effective Sparsification of Neural Networks with Global Sparsity Constraint}

\author{Xiao Zhou$^\text{1}$\thanks{Equal contribution},~~ Weizhong Zhang$^\text{1}$\footnotemark[1], ~~Hang Xu$^\text{2}$, ~~Tong Zhang$^\text{1}$\\
$^1$The Hong Kong University of Science and Technology\qquad$^2$Huawei Noah’s Ark Lab\\
% Institution1 address\\
{\tt\small xzhoubi@connect.ust.hk, weizhong@ust.hk, xu.hang@huawei.com, tongzhang@tongzhang-ml.org}
}
\maketitle
\pagestyle{empty}
\thispagestyle{empty}

%%%%%%%%% ABSTRACT
\begin{abstract}
  Weight pruning is an effective technique to reduce the model size and inference time for deep neural networks in real-world deployments. However, since magnitudes and relative importance of weights are very different for different layers of a neural network, existing methods rely on either manual tuning or handcrafted heuristic rules to find appropriate pruning rates individually for each layer. This approach generally leads to suboptimal performance.  In this paper, by directly working on the probability space, we propose an effective network sparsification method called {\it probabilistic masking} (ProbMask), which solves a natural sparsification formulation under global sparsity constraint. The key idea is to use probability as a global criterion for all layers to measure the weight importance. An appealing feature of ProbMask is that the amounts of weight redundancy can be learned automatically via our constraint and thus we avoid the problem of tuning pruning rates individually for different layers in a network. Extensive experimental results on CIFAR-10/100 and ImageNet demonstrate that our method is  highly effective, and can outperform previous state-of-the-art methods by a significant margin, especially in the high pruning rate situation. Notably, the gap of Top-1 accuracy between our ProbMask and existing methods can be up to 10\%.  As a by-product, we show ProbMask is also highly effective in identifying supermasks, which are subnetworks with high performance in a randomly weighted dense neural network.
\end{abstract}

%%%%%%%%% BODY TEXT
\section{Introduction}
Weight pruning \cite{han2015deep} is a popular technique for alleviating the weight redundancy in deep neural networks (DNNs) to improve inference efficiency and decrease computation demands. Typical pruning algorithms usually prune the unimportant weights by   developing proper criteria. It is repeatedly reported in the literature \cite{guo2016dynamic,liu2018rethinking,zeng2018mlprune,li2016pruning} that by pruning one can reduce the neural network size and improve the inference efficiency significantly with quite slight or even negligible loss on performance, which makes  deploying large-scale DNNs on equipment with limited computational and memory budget possible. 

What can serve as a suitable global comparator to measure weight importance and identify sparsity level for different layers is a 
long-standing problem \cite{gale2019state} though impressive results have been achieved. We know that the core module in pruning is the explicit or implicit  criterion for identifying the redundant weights, and it is difficult to develop a global criterion for the weights in all the layers.  For example, in \cite{han2015deep}, the authors propose a simple yet effective criterion, i.e., for each layer it prunes all the  weights below a certain threshold in a fully trained network. The threshold is obtained by sorting weight by its magnitude and retrieving the weight magnitude at the target pruning rate. The criterion is weight magnitude in this case. Notice that the magnitudes of the weights across layers could be quite different and different layers could have different amount of redundancy. If we use a global threshold for all the layers, then almost all the weights in certain layers could be pruned in order to achieve high enough pruning ratio, which will be verified in Section \ref{sec:ablation-study}. Thus, we need to set a proper threshold or pruning ratio for each layer individually.  In the networks with numerous layers, it is very difficult and even impossible  to find the optimal thresholds or pruning ratios for all the layers manually. One reasonable compromise for such dilemma is to set sparsity level uniformly for different layers. However, this results in imperfect weight allocation obviously and gives unsatisfactory results on high pruning rates. 

\iffalse
Recently \cite{kusupati2020soft} also notice the drawback of uniform weight allocation and gives considerable improvements by learnable sparsity. However, its performance is limited by both the modification of original network pruning problem and the imperfect manual choice of threshold function.
\fi

In this paper, to address the above limitations, we propose an effective network sparsification method called {\it probabilistic masking} (ProbMask). Firstly, we know that network pruning can be naturally formulated into a problem of finding a sparse binary mask $\bs{m}$ as well as the weights at the same time to minimize the empirical loss (\ref{eqn:discrete}). If the component $m_i$ is equal to $0$, it means that the corresponding weight is pruned. However, it is a discrete optimization problem and hard to solve. We notice that if we view the components $\boldsymbol{m}_i$ in the mask  as independent Bernoulli random variables with probability $s_i$ being $1$ and  probability $(1-s_i)$ being $0$ and reparameterize them  w.r.t. its probability, then the loss in problem (\ref{eqn:discrete}) would become continuous over the probability space. Due to the nature of probability, probability can be used as a global criterion in all the layers. Therefore, we can control the model size via forcing the sum of all the probabilities $s_i$ of the mask smaller than a proper value, leading to a global sparsity constraint in the probability space. In this way, the discrete optimization problem (\ref{eqn:discrete}) is transformed into a constrained expected loss minimization problem (\ref{eqn:continuous}) over a probability space, which is continuous. Finally, we adopt the Gumbel-Softmax trick to solve the continuous problem. As the optimizer goes on, the probabilities $s_i$ would converge to either 0 or 1, i.e., $\bs{m}$ would become close to a deterministic sparse mask. Thus, a fully trained mask would have quite low variance, making the loss of the sampled sparse network according to $\boldsymbol{m}$ close to the excepted loss in problem (\ref{eqn:continuous}).  Another appealing feature of our proposed method is that the amount of weight redundancy in each layer can be identified automatically by our global  sparsity constraint and thus we do not need to choose different pruning ratios for different layers.

Experimental results on network pruning and supermask \cite{zhou2019deconstructing} finding demonstrate  that our method is much more effective than the state-of-the-art methods on both small scale datasets and large scale datasets and can outperform them with a significant margin when the pruning rate is high. 

The contribution and novelty of ProbMask can be summarized as follows: 

1) We provide evidence showing that probability can serve as a suitable global comparator to measure weight importance and identify sparsity level for different layers, which is a long-standing problem \cite{gale2019state}.

2) We present a natural formulation of global sparsity constraint, and an optimization method that is practically effective. Our solution fixes the training and testing performance discrepancy problem observed in practice, which led to the failure of previous methods \cite{louizos2017learning} on ImageNet \cite{gale2019state}. 

3) We demonstrate the effectiveness of using probability as global comparator on small-scale and large-scale problems and various models and achieve state-of-the-art results on Top-1 accuracy and accuracy-versus-FLOPS curve.

4) We show ProbMask can also serve as a powerful tool for identifying  supermasks,  which  are  subnetworks  with  high  performance  in  a  randomly  weighted dense neural network, and we achieve state-of-the-art results on Top-1 accuracy on CIFAR-100 under high pruning rates. 

\textbf{Notations:} Let $\|\cdot\|_0, \|\cdot\|_1$ and $\|\cdot\|_2$ be the $\ell_0, \ell_1$ and $\ell_2$ norm of a real valued vector, respectively. We denote $\mathbf{1}\in \mathbb{R}^n$ to be a vector with all components equal to 1.   In addition, $\{0,1\}^n$ is the set of $n$-dimensional vectors with each coordinate valued in $\{0,1\}$.
\iffalse
1. 举例子，不同算法，不同层的prune rate，可能有的算法的结果为0 *
2. 举例子，比如一个图片，我们的算法能够得到好的中间层，其他的不行
3. 对比实验，固定每层prune rate一样 *
4. 对比实验，如果不train weight
5. TA，discrete 
6. score量级不一样，我们的量级是一样的 *因为不同层的量级不一样，所以会导致出现层为0，挑几层画出它们的直方图。 解释1.

7. 为什么gmp uniform能够做好？g

8. 看一下directional pruning的magnitude random pruning比我们的prune by abs结果还要好。 可能是这个原因，他那个random也是逐步降低prune rate得到的结果，不是random选然后再finetune
9. 初始化 概率以weight绝对值的比例
10.给weight 复活的机会，根据概率

\fi

\section{Related Work}
Below, we first review the related work on network pruning. Next we review training methods for obtaining sparse networks which can be divided into two groups: dense-to-sparse training and sparse-to-sparse training. Then we review some probability-based methods for obtaining sparse networks and point out some limitations to differentiate them from our work. Finally we review another line of research on Lottery Tickets Hypothesis, SuperMask and Foresight Pruning.
\subsection{Network Pruning}
Network Pruning \cite{han2015learning, guo2016dynamic,zeng2018mlprune,li2016pruning,luo2017thinet,he2017channel,zhu2017prune, kang2020operation, wang2019structured, renda2020comparing, ye2020good} has been extensively studied in recent years to reduce the model size and improve the inference efficiency of deep neural networks. Since it is a widely-recognized property that modern neural networks are always over-parameterized, pruning methods are developed to remove unimportant parameters in the fully trained dense networks to alleviate such redundancy. According to the  granularity of pruning, existing pruning methods can be roughly divided into two categories, i.e., unstructured pruning and structured pruning. The former one is also called weight pruning, which removes the unimportant parameters in an unstructured way, that is, any element in the weight tensor could be removed. The latter one removes all the weights in a certain group together, such as kernel and filter. Since structure is taken into account in pruning, the pruned networks obtained by structured pruning are available for efficient inference on standard computation devices. In both structured and unstructured pruning methods, their key idea is to propose a proper implicit or explicit criterion (e.g., magnitude of the weight \cite{han2015deep, han2015learning, guo2016dynamic, zhu2017prune, frankle2018lottery, mostafa2019parameter, bellec2017deep, mocanu2018scalable, wortsman2019discovering}, scores based on Hessian, momentum or gradient \cite{menickrigging, lee2018snip, zeng2018mlprune, lecun1990optimal, hassibi1993second, dettmers2019sparse}) to evaluate the importance of the weight, kernel or filter and then remove the unimportant ones. The results in the literature \cite{guo2016dynamic,liu2018rethinking,zeng2018mlprune,li2016pruning, renda2020comparing, kusupati2020soft, menickrigging, wortsman2019discovering} demonstrate that pruning methods can significantly improve the inference efficiency of DNNs with minimal performance degradation, making the deployment of modern neural networks on resource-limited devices possible.

\subsection{Dense-to-sparse and Sparse-to-sparse Training}

We follow the convention of \cite{kusupati2020soft} to divide training algorithms for obtaining sparse networks into two groups: dense-to-sparse training and sparse-to-sparse training. Dense-to-sparse training starts with a dense network and obtains a sparse network at the end of the training \cite{han2015learning, zhu2017prune, molchanov2017variational, frankle2018lottery, renda2020comparing, xiao2019autoprune, srinivas2017training, louizos2017learning, wortsman2019discovering}. ProbMask belongs to the group of dense-to-sparse training. \cite{han2015deep, zhu2017prune, frankle2018lottery, renda2020comparing} follows the idea of using weight magnitude as the criterion. \cite{zhu2017prune} manually set a uniform sparsity budget for different layers. \cite{renda2020comparing} achieves strong results but needs multiple rounds of pruning and retraining. \cite{xiao2019autoprune} assigns auxiliary scores to weights and use it as the criterion. \cite{xiao2019autoprune} suffers from the bias induced by the approximation of the step function and will have gradient vanishing problem when using ReLU and SoftPlus as the approximator. This makes the auxiliary scores hard to act as a global criterion. \cite{srinivas2017training, louizos2017learning, molchanov2017variational} base its criterion on reparameterization of probability and have the most connections with our work. We will fully discuss them in the next subsection. 

Sparse-to-sparse training starts with a sparse network and maintain the sparsity during training \cite{bellec2017deep, mocanu2018scalable, mostafa2019parameter, menickrigging, dettmers2019sparse}. It uses criterion like weight magnitude, weight gradient magnitude, momentum of weight to reallocate sparsity through training. Conceptually, sparse-to-sparse training can  reduce the computational cost during training but it is hard to take effect without the support of sparse convolution framework on GPU. The performance of sparse-to-sparse training generally falls behind dense-to-sparse training under the same setting as shown in \cite{wang2020picking, kusupati2020soft}.

\subsection{Probability-based Methods}
Compared to directly treating probability as the trainable variable, \cite{srinivas2017training, louizos2017learning, molchanov2017variational} consider probability as the hidden state and optimize on another space by reparameterization of probability. \cite{srinivas2017training, louizos2017learning, molchanov2017variational} achieve strong empirical results while we observe several shortcomings in the reparameterization process. \cite{srinivas2017training} approximates the gradient through sampling process by biased STE (straight-through estimator) \cite{bengio2013estimating} and represent probability as the output of a hard-sigmoid function which induces gradient vanishing. \cite{louizos2017learning} reparameterizes w.r.t hard-concrete and \cite{louizos2017learning} is reported to fail to work on ImageNet dataset because of the performance gap between training and testing phases \cite{gale2019state}. \cite{molchanov2017variational} also exhibits gradient vanishing problem due to function pattern of KL divergence and performance gap between training and testing phases by generating the test model by a cut-off manner rather than sampling. We solve the aforementioned problems by directly optimizing over probability space. We empirically demonstrate that probabilities finally converge to either 0 or 1 after training in Section \ref{sec:ablation-study}, leading to a binary mask and fixing the training and testing performance discrepancy problem observed in practice. Besides, by explicilty control the model size via global sparsity constraint, users don't need to tune regularization parameters to achieve desired model size, which is a missing feature in \cite{srinivas2017training, louizos2017learning, molchanov2017variational}.

\subsection{Lottery Tickets Hypothesis, Supermask and Foresight Pruning}

Lottery Ticket Hypothesis was proposed in \cite{frankle2018lottery}, which conjectures and verifies that there exists sparse subnetworks which can be trained directly to achieve even better performance than dense counterparts with less training time. \cite{zhou2019deconstructing} further analyzes the conditions for such phenomenon to hold and propose supermask, which conveys an intriguing idea that a good mask is enough to achieve surprisingly good performance with randomly initialized weights. \cite{ramanujan2020s} further asks the question what's hidden in a randomly weighted neural network and proposes a more effective algorithm on finding supermasks. Lottery Ticket Hypothesis also sheds light on whether we could find such subnetwork without training a dense network. \cite{lee2018snip} propose the first algorithm to find such subnetworks and  \cite{wang2020picking} further improves performance on high pruning rates. However, both of them could not achieve better performance than the latest pruning methods.

% The key idea of LTH \citep{frankle2018lottery}  concentrates on whether an already sparsified network with desired initializations suffices to obtain desired compressed network with comparable performances with dense counterparts. \citet{frankle2019stabilizing} shows that weight status after thousands steps of training of dense networks are needed in the large network situation, e.g. ResNet-50. 

% SNIP\citep{lee2018snip} explores the first algorithm to find sparsified networks at initialization to obtain good final performance after training by preserving the loss of the original randomly initialized network. GraSP\citep{wang2020picking} develops another foresight pruning criterion based on preserving the gradient flow to obtain better performance on high pruning rates.
\section{Effective Sparsification with Global Sparsity Constraint}
Below, we present our proposed network sparsification framework and the method for solving  the minimization problem in the framework.

\subsection{A Probabilistic  Sparsification Framework}

Let $\mathcal{D}$ be a dataset consisting of $N$ i.i.d. samples $\left\{\left(\mathbf{x}_{1}, \mathbf{y}_{1}\right), \ldots,\left(\mathbf{x}_{N}, \mathbf{y}_{N}\right)\right\}$, $\boldsymbol{w}\in \mathbb{R}^n$ be the weights of a neural network. We denote $\boldsymbol{m}\in \{0,1\}^n$ to be the masks of the weights. $m_i=0$ means the weight $w_i$ is pruned and otherwise $w_i$ is kept.  The problem of training sparse neural networks can be naturally  formulated into the following empirical risk minimization problem: 
\begin{gather}
\min_{\boldsymbol{w}, \boldsymbol{m}} ~\mathcal{L}(\boldsymbol{w}, \boldsymbol{m}):=\frac{1}{N}\sum_{i=1}^{N} \ell\left(h\left(\mathbf{x}_{i} ; \boldsymbol{w\circ m}\right), \mathbf{y}_{i}\right) \label{eqn:discrete}\\
s.t.~ \boldsymbol{w}\in \mathbb{R}^n, \norm{\boldsymbol{m}}_1 \leq K \mbox{ and }  \boldsymbol{m}\in \{0,1\}^n \mbox, \nonumber 
\end{gather}
where $h(\cdot ; \boldsymbol{w\circ m})$ is output of the pruned  network with $\circ$ being the element-wise product of two vectors, and $\ell(\cdot, \cdot)$ is the loss function, e.g, the squared loss for regression or cross entropy loss for classification.  $K = kn$ is the model size we want to reduce the network to, i.e., the number of remaining weights after pruning and $k$ is the remaining ratio of model weights. In this framework, the model size is controlled by a single constraint which avoids tuning the pruning rate for each layer. However, since the objective is discrete with respect to the  mask $\boldsymbol{m}$, problem (\ref{eqn:discrete}) is hard to solve and thus cannot be applied in practice.

We notice that if  we view each component of mask $\boldsymbol{m}$ as a binary random variable and reprameterize problem (\ref{eqn:discrete}) with respect to the distributions of this random variable, then problem (\ref{eqn:discrete}) can be relaxed into an excepted loss minimization problem over the weight and probability spaces, which is continuous. We need to point out that this is a very tight relaxation since empirical observations show that probabilities $s_i$ converge to 0 or 1 after training (Section \ref{sec:ablation-study}). Specifically, we can view $m_i$ as a Bernoulli random variable with probability $s_i$ to be $1$ and $1-s_i$ to be $0$, that is $m_i \sim \operatorname{Bern}(s_i)$, where $s_i \in [0,1]$. Assuming the variables $m_i$ are independent, then we can get the distribution function of $\boldsymbol{m}$, i.e.,  $p(\boldsymbol{m}|\boldsymbol{s}) = \Pi_{i=1}^{n} (s_i)^{m_i}(1-s_i)^{(1-m_i)}$. Thus, the model size can be controlled by the sum of the probabilities $s_i$, i.e.,  $\boldsymbol{1}^\top\boldsymbol{s}$, since $\mathbb{E}_{\boldsymbol{m}\sim p(\boldsymbol{m}|\boldsymbol{s})}\|\boldsymbol{m}\|_0 = \sum_{i=1}^n s_i$. Then the discrete constraint $\|\boldsymbol{m}\|_1\leq K $ in problem (\ref{eqn:discrete}) can be transformed into $\boldsymbol{1}^\top \boldsymbol{s} \leq K$ with each $s_i\in [0,1]$, which is continuous and convex. Therefore, problem (\ref{eqn:discrete}) can be relaxed  into the following excepted loss minimization problem: 
% \begin{center}
% \begin{equation}
% \min_{\boldsymbol{w}, \boldsymbol{s}} \displaystyle ~\mathbb{E}_{ p(\boldsymbol{m}|\boldsymbol{s})} \mathcal{L}(\boldsymbol{w}, \boldsymbol{m}) \label{eqn:continuous} \\ 
% s.t. ~\boldsymbol{w}\in \mathbb{R}^n, \boldsymbol{1}^\top \boldsymbol{s}  \leq K \mbox{ and } \boldsymbol{s}\in [0,1]^n \nonumber
% \end{equation}
% \end{center}
\begin{gather}
\min_{\boldsymbol{w}, \boldsymbol{s}} \displaystyle ~\mathbb{E}_{ p(\boldsymbol{m}|\boldsymbol{s})} ~\mathcal{L}(\boldsymbol{w}, \boldsymbol{m}) \label{eqn:continuous} \\ 
s.t. ~\boldsymbol{w}\in \mathbb{R}^n, \boldsymbol{1}^\top \boldsymbol{s}  \leq K \mbox{ and } \boldsymbol{s}\in [0,1]^n. \nonumber
\end{gather} 

% \begin{equation}
% \begin{gathered}
% \min_{\boldsymbol{w}, \boldsymbol{s}} \displaystyle ~\mathbb{E}_{ p(\boldsymbol{m}|\boldsymbol{s})} \mathcal{L}(\boldsymbol{w}, \boldsymbol{m}) \label{eqn:continuous} \\ 
% s.t. ~\boldsymbol{w}\in \mathbb{R}^n, \boldsymbol{1}^\top \boldsymbol{s}  \leq K \mbox{ and } \boldsymbol{s}\in [0,1]^n
% \end{gathered}
% \end{equation}

%{\textcolor[rgb]{1,0,0}{Readers may ask why we don't formulate the task in the logit space (probability is generated by sigmoid transformation of logit.) Acutually, formulating the problem with logit introduces sigmoid function in the gradient calculation process, thus casuing the well-known problem of gradient vanishing. This induces accuracy loss experimentally.}}

\textbf{Discussion.} Appealing features of ProbMask:
\begin{itemize}
    \item The constraints in problem (\ref{eqn:continuous}) can be rewritten as $\|\boldsymbol{s}\|_1 \leq K $ and $\boldsymbol{s}\in [0,1]^n$. Due to  this $\ell_1$ norm constraint, the optimal $\boldsymbol{s}$ is sparse. Most of $s_i$ would be either $0$ or $1$ with  high probability, making $\boldsymbol{m}$ converge to a deterministic mask. Therefore, $\boldsymbol{s}$ after training would have a quite low variance and thus the loss of a randomly sampled $\boldsymbol{m}$ would be close to the expected loss in Eqn.(\ref{eqn:continuous}).
    \item  Compared with problem (\ref{eqn:discrete}), problem (\ref{eqn:continuous}) is continuous. Moreover, the feasible region  of problem (\ref{eqn:continuous}) is quite simple, which is actually the intersection of the cube $[0,1]^n$ and the half space $\mathbf{1}^\top \boldsymbol{s}\leq K$. For such simple set, the projection operator has an explicit expression, please see Theorem \ref{theorem 1} for the details. This makes it possible to adopt the efficient optimization algorithms such as projected gradient descent to solve problem (\ref{eqn:continuous}).
    \item %In most of the existing pruning methods, one has  to set different  pruning ratios for different layers in a  network  manually or using handcrafted hard rules. The reason is that the weights in different layers have quite different magnitudes of importance under their proposed criteria. Therefore, it is unreasonable to set a global threshold for all the layers to determine whether to prune a certain weight.  Moreover, since different layers always have different amounts of redundancy, it is very difficult and even impossible to find the optimal pruning ratios for each layer. This limitation hinders the existing pruning methods from achieving higher pruning ratios, which will be verified in the experiment section (see Section \ref{sec:ablation-study}). 
    In our framework, the problem is reparameterized with respect to probability, which can be used as a global criterion to measure the importance of the weights in different layers. Note that the constraint is applied over all probability for different layers, rather than setting a uniform sparsity across layers. 
    The amount of redundancy in each layer of the neural network can be learned automatically in the process of solving problem (\ref{eqn:continuous}), which will be verified in Section \ref{sec:ablation-study}. Therefore, we avoid setting pruning ratio for each layer manually. The benefits of globally comparable property of probability on the model size and accuracy will be further verified in Section \ref{sec:exp}. 
    
    % We will show in Section \ref{sec:ablation-study} that ProMask automatically learns a wise balance between uniform budget and cutting one layer off by globally uncomparable criteria. 
  
    %We obtain an unified representation of the importance of weights across layers---probability, overcoming the problem of different importance magnitudes across layers. Comparison of importance between weights from different layers become straight forward in this senario. Prune rate does not need manual setting for different layers and thus global pruning ratio constraint is applied in our algorithm, leading to better minimum theoretically.
\end{itemize}

%  All in all, \cite{louizos2017learning} is not based on probability but just use reparameterization to give inaccurate results.

% \cite{srinivas2017training} solves the expected loss minimization using straight-through estimator \cite{bengio2013estimating} and ensures the convergence of probabilities to 0 or 1 through two regularizers. However it includes much bias produced by ignoring the Heaviside in the likelihood during the gradient evaluation \cite{louizos2017learning}. Besides, a pretrained dense model is mandatory for it to ensure good performance and this enduces extra cost. Both of them suffer from implicit relationship between regularizer parameters and desired prune rate and much efforts are needed for searching.

% There also exists method sharing similar ideas from the area of Neural Architecture Search(NAS). SNAS \cite{xie2018snas} also attempts to solve an expected loss minimization problem while searches solutions in the logit space.

%概率是统一的，所以我们不需要对每一层设置prune rate，理论上找一个更优的解。
\begin{tiny}
\begin{algorithm*}[htb!]
\caption{Probabilistic Masking (ProbMask)}
\label{alg:SST}
\begin{algorithmic}[1]
\REQUIRE target remaining ratio $k_f$, a dense network $\bs{w}$.
\STATE Initialize $\bs{w}$, assign probabilities $\bs{s}$ to weights $\bs{w}$, let $\bs{s}=\mathbf{1}$ and  $\tau=k= 1$.
\FOR {training epoch $t = 1, 2 \ldots T$}
\STATE Decrease the temperature annealing parameter by  $\tau = 0.97(1-t/T)+0.03$.
\STATE Update $k$ according to Eqn.(\ref{eqn:update-k}).
\FOR {each training iteration}
\STATE Sample mini batch of data $\mathcal{B} = \left\{\left(\mathbf{x}_{1}, \mathbf{y}_{1}\right), \ldots,\left(\mathbf{x}_{B}, \mathbf{y}_{B}\right)\right\}$.
\STATE Generate  $\boldsymbol{g_1}^{(i)}$ and $\boldsymbol{g_0}^{(i)}$ with each element sampled from Gumbel(0, 1), $ i=1,2,\dots,I$.
\STATE  $\bs{s} \leftarrow \operatorname{proj}_C(\bs{z}), \mbox{with }\bs{z} = \bs{s} -  \eta \frac{1}{I} \sum_{i=1}^I\nabla_{\boldsymbol{s}}\mathcal{L}_{\mathcal{B}}\Big(\boldsymbol{w}, \sigma\big(\frac{\log(\frac{\boldsymbol{s}}{\boldsymbol{1}-\boldsymbol{s}})+\boldsymbol{g_1}^{(i)}-\boldsymbol{g_0}^{(i)}}{\tau}\big)\Big).$ 
%$\mathcal{L}_{\mathcal{B}}$ is the loss calculated over minibatch $\mathcal{B}$.
\STATE $\bs{w} \leftarrow \bs{w} -  \eta \frac{1}{I} \sum_{i=1}^I\nabla_{\boldsymbol{w}} \mathcal{L}_{\mathcal{B}}\Big(\boldsymbol{w}, \sigma\big(\frac{\log(\frac{\boldsymbol{s}}{\boldsymbol{1}-\boldsymbol{s}})+\boldsymbol{g_1}^{(i)}-\boldsymbol{g_0}^{(i)})}{\tau}\big)\Big)$
\ENDFOR
\ENDFOR

%\ENSURE Sample one mask $\bs{m}$ from $\bs{s}$ and  obtain the pruned network with $\bs{w}\circ\bs{m}$. Return the pruned network $\bs{w}\circ \boldsymbol{m}$ by sampling a mask $\bs{m}$ from the distribution $p(\boldsymbol{m}|\boldsymbol{s})$.
\RETURN A pruned network $\bs{w}\circ \boldsymbol{m}$ by sampling a mask $\bs{m}$ from the distribution $p(\boldsymbol{m}|\boldsymbol{s})$.
\end{algorithmic}
\end{algorithm*}
\end{tiny}

\subsection{Optimization with Projected Gradient Descent} \label{opt}

% We do not reparameterize problem (\ref{eqn:discrete}) in the logit space (probability is generated by sigmoid transformation of logit), since we need a  {\it simple feasible region}, in a sense that the projection on this region  can be solved very efficiently. Another benefit of directly optimizing on the probability space is that we can avoid gradient vanishing induced by sigmoid transformation.

Below, we present our training method for  problem (\ref{eqn:continuous}). We update both the weights $\boldsymbol{w}$ and the probability $\boldsymbol{s}$ at training time. At test time, we obtain the sparse network $\bs{w}\circ\boldsymbol{s}$ by sampling according to probability $\boldsymbol{s}$. We adopt projected gradient descent(PGD) as the optimizer and the details are as follows.

\textbf{[Gradient Computation]} The difficulty lies in computing the gradient of the expected loss with respect to the probability. Therefore, in this paper, we adopt Gumbel-Softmax \cite{jang2016categorical, maddison2016concrete} trick to calculate the gradient, with which  the gradient w.r.t. weights and probability can be calculated in the following form:
\begin{align}
    &\nabla_{\boldsymbol{s}, \boldsymbol{w}} \E_{p(\boldsymbol{m}|\boldsymbol{s})} \mathcal{L}\left(\boldsymbol{w}, \boldsymbol{m}\right)\nonumber \\ 
    =& \E_{\boldsymbol{g_0}, \boldsymbol{g_1}} \nabla_{\boldsymbol{s}, \boldsymbol{w}} \mathcal{L}\Big(\boldsymbol{w}, \mathds{1}\big(\log(\frac{\boldsymbol{s}}{\boldsymbol{1}-\boldsymbol{s}})+\boldsymbol{g_1}-\boldsymbol{g_0} \geq 0\big)\Big) \label{for proof}\\
    \approx& \E_{\boldsymbol{g_0}, \boldsymbol{g_1}} \nabla_{\boldsymbol{s}, \boldsymbol{w}} \mathcal{L}\Big(\boldsymbol{w}, \sigma\big(\frac{\log(\frac{\boldsymbol{s}}{\boldsymbol{1}-\boldsymbol{s}})+\boldsymbol{g_1}-\boldsymbol{g_0}}{\tau}\big)\Big) \label{before sample}\\
    \approx & \frac{1}{I} \sum_{i=1}^I\nabla_{\boldsymbol{s}, \boldsymbol{w}} \mathcal{L}\Big(\boldsymbol{w}, \sigma\big(\frac{\log(\frac{\boldsymbol{s}}{\boldsymbol{1}-\boldsymbol{s}})+\boldsymbol{g_1}^{(i)}-\boldsymbol{g_0}^{(i)}}{\tau}\big)\Big), \label{after sample} 
\end{align}
where  $\mathds{1}(\boldsymbol{A})\in \{0,1\}^n$ is the indicator function. $\boldsymbol{g_0}$ and $\boldsymbol{g_1}$ are two random variables in $\mathbb{R}^n$, with each element i.i.d sampled from $\operatorname{Gumbel}(0,1)$ distribution. $\boldsymbol{g_1}^{(i)}$ and $\boldsymbol{g_0}^{(i)}$ with $ i=1,2,\dots,I$ are $2I$ sampled instances. $\sigma(\cdot): \mathbb{R}^n \rightarrow (0,1)^n$ here is the  element-wise sigmoid function, i.e., $\sigma(\boldsymbol{x}) = \frac{1}{1+e^{-\boldsymbol{x}}}$ for any $\boldsymbol{x}\in \mathbb{R}^n$. $\tau$ is a  temperature annealing parameter decreasing linearly during training and precise choice of the decreasing function contributes to convergence of probability to a deterministic state. We will present empirical obersavtions in Section \ref{sec:ablation-study} and provide some informal insights on such contribution from precise choice of temperature decreasing function in appendix. From Eqn.(\ref{before sample}) to Eqn.(\ref{after sample}), multiple networks are sampled to obtain a steady gradient flow with a low variance. The proof of the equations above is placed in appendix.
%The Term $\mathds{1}(\log(\boldsymbol{s})-\log(\boldsymbol{1}-\boldsymbol{s})+\boldsymbol{g_1}-\boldsymbol{g_0} \geq 0)$ can be seen as an equivalent transformation of mask $\boldsymbol{m}$. The sigmoid function $\sigma(\boldsymbol{x})$ together with temperature annealing is applied to obtain an differentiable approximation to the indicator function. 

\textbf{[Projected Gradient Descent]} We denote the feasible region of probability in problem (\ref{eqn:continuous}) as $C$, that is $C = \{\bs{s} \mid \norm{\boldsymbol{s}}_1 \leq K \mbox{ and }\boldsymbol{s}\in [0,1]^n\}$. The theorem below shows that the projection of a vector onto $C$ can be calculated efficiently, which makes PGD applicable. 
\begin{theorem}\label{theorem 1}
For each vector $\bs{z}$, its projection $\bs{s}$ in the set $C$ can be calculated as follows:
\begin{align}
\bs{s} = \min (1, \max(0, \bs{z}-v_{2}^{*}\mathbf{1})).
\end{align}
where $v_{2}^{*} = \max(0, v_{1}^{*})$ with $v_{1}^{*}$ being the solution of the following equation 
\begin{align}
\bs{1}^\top[\min (1, \max(0, \bs{z}-v_{1}^{*}\mathbf{1}))] - K = 0. \label{eqn:theorem1}
\end{align}
\end{theorem}
The equation (\ref{eqn:theorem1}) can be solved by bisection method efficiently. Now we can apply PGD to solve problem (\ref{eqn:continuous}) directly on probability space with explicit sparsity constraint. We provide a complete view of ProbMask in Algorithm \ref{alg:SST}, and supplementary messages can be found in appendix.

\textbf{[Gradually Increasing Pruning Rate]} %ProbMask gradually increases the pruning rate to make a smooth transformation from dense to extremely sparse status. The remaining ratio schedule follows the practice of GMP \cite{zhu2017prune}. 
We increase the pruning rate gradually to make a smooth transformation from dense to sparse status. We ultilize the increasing function of \cite{zhu2017prune},
\begin{align} 
 k=k_{f}+\left(1-k_{f}\right)\left(1-\frac{t-t_{1}}{t_{2}-t_{1}}\right)^{3}, \label{eqn:update-k}
\end{align} 
where $t \in \left\{t_{1}, t_{1}+1, \ldots, t_{2}\right\}$ is the current epoch number and $k_f$ is the targeted remaining ratio. $k$ keeps 1 before epoch $t_{1}$ and  $k_f$ after epoch $t_{2}$. 
\begin{remark}
ProbMask directly works on  the probability space without any further reparameterization, avoiding the drawback of gradient vanishing \cite{srinivas2017training,molchanov2017variational}. Together with the global sparsity constraint, ProbMask finally learns a deterministic state of probability, resulting in a  little performance gap in testing and training phases.
\end{remark}
\begin{remark}
ProbMask can be trained with randomly initialized weights or from pretrained weights. ProbMask can explicitly control the sparsity by choosing a proper $K$ to achieve a desired model size and does not need to search any parameters. 
\end{remark}
\iffalse
\textbf{Remark.}  
\begin{itemize}
    \item ProbMask directly optimizes probability without further reparameterization, avoiding the drawback of gradient vanishing \cite{srinivas2017training, molchanov2017variational}. 
    \item Probabilistic Masking finally learns a deterministic state of probability, resulting in little gap in test time and training time performance \cite{molchanov2017variational, louizos2017learning}. 
    \item Probabilistic Masking can be trained with randomly initialized weights or pretrained weights, and we can directly optimize with target sparsity constraint with pretrained weights.
  
    %We obtain an unified representation of the importance of weights across layers---probability, overcoming the problem of different importance magnitudes across layers. Comparison of importance between weights from different layers become straight forward in this senario. Prune rate does not need manual setting for different layers and thus global pruning ratio constraint is applied in our algorithm, leading to better minimum theoretically.
\end{itemize}
\fi

%For your reminder, $K = k\norm{\bs{m}}_0$. Details and motivations for this function can be found at \citet{zhu2017prune} and specific numbers can be found at Section \ref{expsettings}.

\section{Experiment}\label{sec:exp}

% Now we validate the effectiveness and properties of ProbMask through various experiments. Firstly, we conduct small scale experiments on CIFAR-10/100(\cite{krizhevsky2009learning}) on modern architectures VGG19(\cite{simonyan2014very}) and  ResNet32(\cite{he2016deep}) comparing against two strong network compression baselines, PBW and MLPrune. Next we conduct large scale experiments on ImageNet(\cite{deng2009imagenet}) on ResNet50(\cite{he2016deep}) to showcase the perfomance of ProbMask. Various dynamic sparse training methods(RIGL, STR, DNW, GMP) are included. Besides We study the difference of uniform and non-uniform sparsity budget through experiments on very sparse regime of ResNet32 on CIFAR10. Finally, we study the globally comparable property of probability against the globally non-comparable of importance scores of MLPrune.%(写好一点？)

In this section, we conduct a series of experiments to evaluate the performance of our proposed method. We divide the experiments into two parts. In part one, we conduct lots of relatively small-scaled experiments on CIFAR-10/100 datasets with modern architectures VGG19 \cite{simonyan2014very} and ResNet32 \cite{he2016deep} to verify some appealing properties of our method. % We will verify the claims we made using these results in Section \ref{sec:ablation-study}. 
 In part two, we verify the superiority of our method over state-of-the-art methods by conducting experiments on ImageNet \cite{deng2009imagenet}.  We choose six representative methods PBW (Pruning by Weight, \cite{han2015learning}), MLPrune \cite{zeng2018mlprune}, RIGL \cite{menickrigging}, STR \cite{kusupati2020soft}, DNW\cite{maddison2016concrete}, GMP \cite{zhu2017prune}) as baselines. PBW \cite{han2015learning} is a classic magnitude-based pruning method. MLPrune \cite{zeng2018mlprune} is a latest Hessian-based pruning method showing overall better performance \cite{wang2020picking} against various sparse-to-sparse training methods (SET \cite{mocanu2018scalable}, DEEPR \cite{bellec2017deep}, DSR \cite{mostafa2019parameter}), so we compare with these sparse-to-sparse training methods implicitly in CIFAR experiments. DNW \cite{wortsman2019discovering}, GMP \cite{zhu2017prune}, STR \cite{kusupati2020soft} are state-of-the-art methods on dense-to-sparse training. RIGL \cite{menickrigging} is the state-of-the-art  sparse-to-sparse training method. Due to the space limitation, we postpone the experimental configurations and MobileNetV1 \cite{howard2017mobilenets} experiments into appendix.

\subsection{VGG19 and ResNet32 on CIFAR-10/100 }
  \iffalse
  For PBW and MLPrune, the train-prune-finetune schedule follows the experimental settings in  \cite{liu2018rethinking}. For our ProbMask, we use Adam (\cite{kingma2014adam}) and SGD to update the probabilities and the weights, respectively.  We adopt the  cosine learning rate schedule with the initial learning rate being 6e-3 for probability and 1e-1 for weights. Both ProbMask and STR are trained with 300 epochs and their batch sizes are all set to 256. All hyperparameters like learning rate for weights, are set the same as previous works \cite{ramanujan2020s, liu2018rethinking}. We only tune the learning rate for probability on CIFAR-10 and Conv4 using grid search and directly apply it to CIFAR-100, ImageNet and other networks. Weight decay for Adam is set 0 as default.
  \fi

\begin{figure}[htb!]
\begin{center}
\centering  
\includegraphics[scale=0.4]{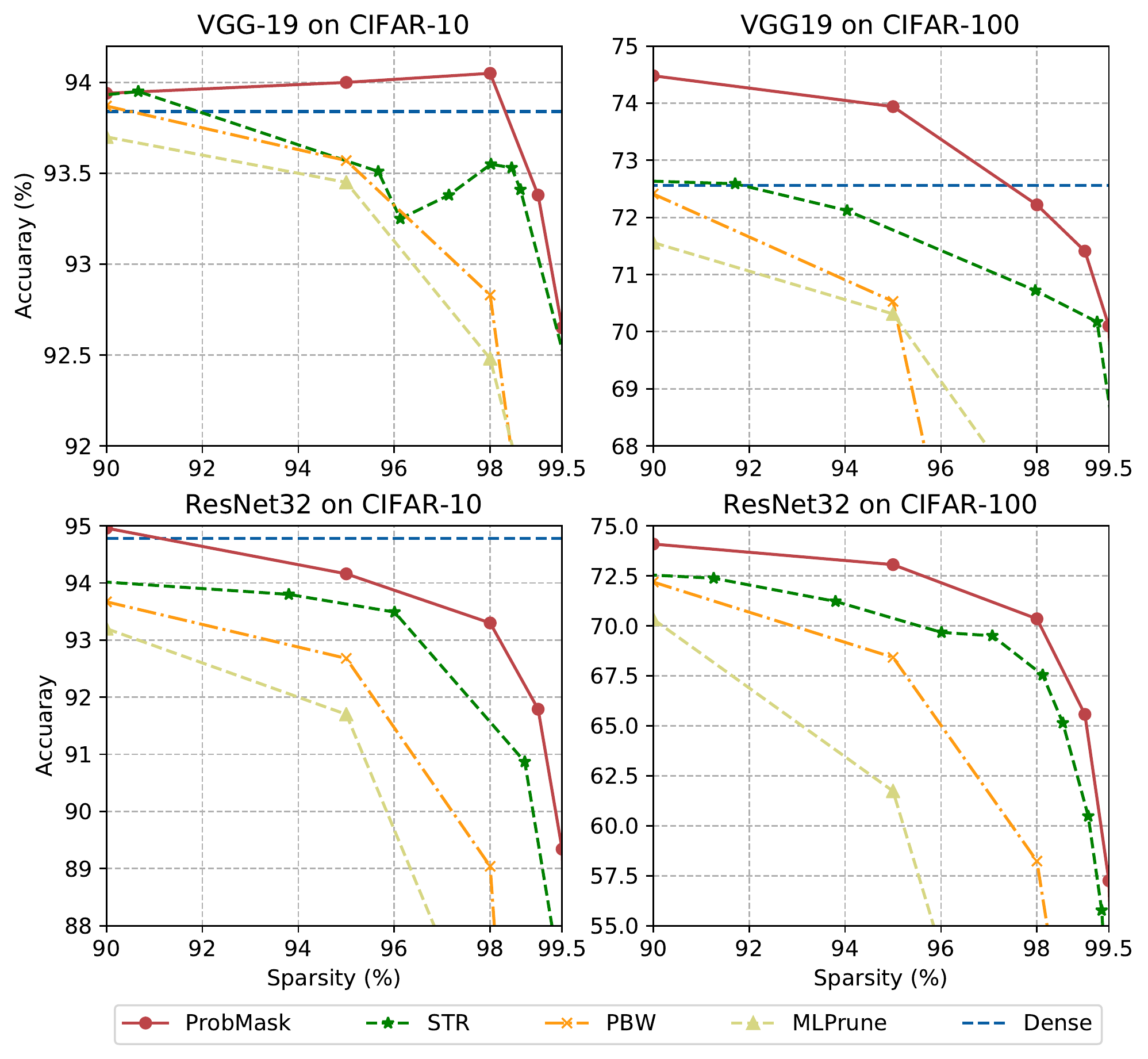}
\end{center}
\caption{Comparison of Top-1 Accuracy on CIFAR-10/100.}\label{fig:result-CIFAR-10-100}
\end{figure}

Table \ref{tab:VGG19-CIFAR10-100} presents the detailed accuracy of PBW, MLPrune and ProbMask  at different pruning ratios. It is very hard to accurately tune weight decay parameter in STR to obtain the desired pruning ratio. Therefore we tune the weight decay parameter  manually to make it have roughly the same pruning ratio range with ProbMask, i.e., 90\% to 99.9\%.% and then  plot the results of ProbMask and STR in Figure \ref{fig:result-CIFAR-10-100} to compare them over the same pruning ratio ranges but at different specific pruning ratios.
% \vspace*{-20pt}

% \vspace*{-20pt}
The results in both Table \ref{tab:VGG19-CIFAR10-100} and Figure \ref{fig:result-CIFAR-10-100} demonstrate that our ProbMask can steadily outperform  the baselines and the superiority  becomes more significant at higher pruning ratios. From Table \ref{tab:VGG19-CIFAR10-100}, we can see that when prune rate come to 99.5\% or higher on CIFAR-10/100, PBW and MLPrune would seriously degrade or even collapse, while our ProbMask can still achieve  significantly  higher.   Figure 1 shows that on CIFAR-100 with VGG19, the gap between ProbMask and STR would be roughly 2\% on average when the remaining ratio is in the range of $[0.9,0.98]$. Pruning ResNet32 is more challenging since VGG19 has about 10 times parameters than ResNet32. In this case, the  gap becomes more significant especially at high pruning ratios, which can be up to 5\% on CIFAR-100 experiments. The superiority of ProbMask over such  high prune ratios attributes to our global sparsity constraint, allowing us to have non-uniform sparsity budgets across layers. This will be further validated in the  ablation study in Section \ref{sec:ablation-study}. 

\begin{table*}[htb!]
\begin{center}
{\footnotesize
\begin{tabular}{p{1.9cm}<{\centering} p{0.8cm}<{\centering}p{0.8cm}<{\centering}p{0.8cm}<{\centering}p{0.8cm}<{\centering}p{0.8cm}<{\centering}p{0.85cm}<{\centering}p{0.8cm}<{\centering}p{0.8cm}<{\centering}p{0.8cm}<{\centering}p{0.8cm}<{\centering}p{0.8cm}<{\centering}p{0.85cm}<{\centering}}
\toprule
Dataset& & &\multicolumn{2}{c}{CIFAR-10}&&& &&\multicolumn{2}{c}{CIFAR-100}&& \\ \cmidrule(){1-13}
Ratio& 90\% & 95\% & 98\% & 99\% & 99.5\% & 99.9\% & 90\% & 95\% & 98\% & 99\% & 99.5\% & 99.9\% \\ \cmidrule(r){1-1} \cmidrule(l){2-7} \cmidrule(l){8-13}
VGG19& 93.84 & - & - & - & - & -&72.56 & - & - & - & - & - \\  \cmidrule(r){1-1} \cmidrule(l){2-7} \cmidrule(l){8-13}
PBW \cite{han2015deep}& 93.87& 93.57& 92.83& 90.89 & 10.00& 10.00 & 72.41& 70.53& 58.91 & 1.00& 1.00 & 1.00 \\  \cmidrule(r){1-1} \cmidrule(l){2-7} \cmidrule(l){8-13}
MLPrune \cite{zeng2018mlprune}& 93.70& 93.45 & 92.48& 91.44& 88.18 &65.38 & 71.56 & 70.31 & 66.77 & 60.10 & 50.98 & 5.58\\  \cmidrule(r){1-1} \cmidrule(l){2-7} \cmidrule(l){8-13}
ProbMask &\textbf{93.94} &\textbf{94.00}& \textbf{94.05}& \textbf{93.38} & \textbf{92.65}& \textbf{89.79} & \textbf{74.48}& \textbf{73.94} & \textbf{72.22} & \textbf{71.41} & \textbf{70.10} & \textbf{60.41} \\
\midrule[0.6pt]
% \cmidrule(){1-13}
ResNet32 & 94.78 & - & - & - & - & - &75.94 & - & - & - & - & -\\  \cmidrule(r){1-1} \cmidrule(l){2-7} \cmidrule(l){8-13}
PBW \cite{han2015deep}& 93.67& 92.68& 89.04& 77.03&73.03 & 38.64 & 72.19 & 68.42 & 58.23 & 43.00 & 20.75 & 5.96 \\  \cmidrule(r){1-1} \cmidrule(l){2-7} \cmidrule(l){8-13}
MLPrune \cite{zeng2018mlprune}& 93.20& 91.70 & 85.64& 76.88& 67.66 &36.09 & 70.33 & 61.73 & 37.86 & 22.38 & 13.85 & 5.50\\  \cmidrule(r){1-1} \cmidrule(l){2-7} \cmidrule(l){8-13}
ProbMask &\textbf{94.96} &\textbf{94.16}& \textbf{93.30}& \textbf{91.79} & \textbf{89.34}& \textbf{76.87} & \textbf{74.09}& \textbf{73.06} & \textbf{70.35} & \textbf{65.57} & \textbf{57.25} & \textbf{26.72} \\  \bottomrule
\end{tabular}
}
\end{center}
\caption{Accuracy of VGG19 and ResNet32   on CIFAR-10/100 at different pruning ratios.}\label{tab:VGG19-CIFAR10-100}
\end{table*}

\subsection{ResNet50 on ImageNet-1K}
\begin{figure}[H]
\centering  
\includegraphics[scale=0.281]{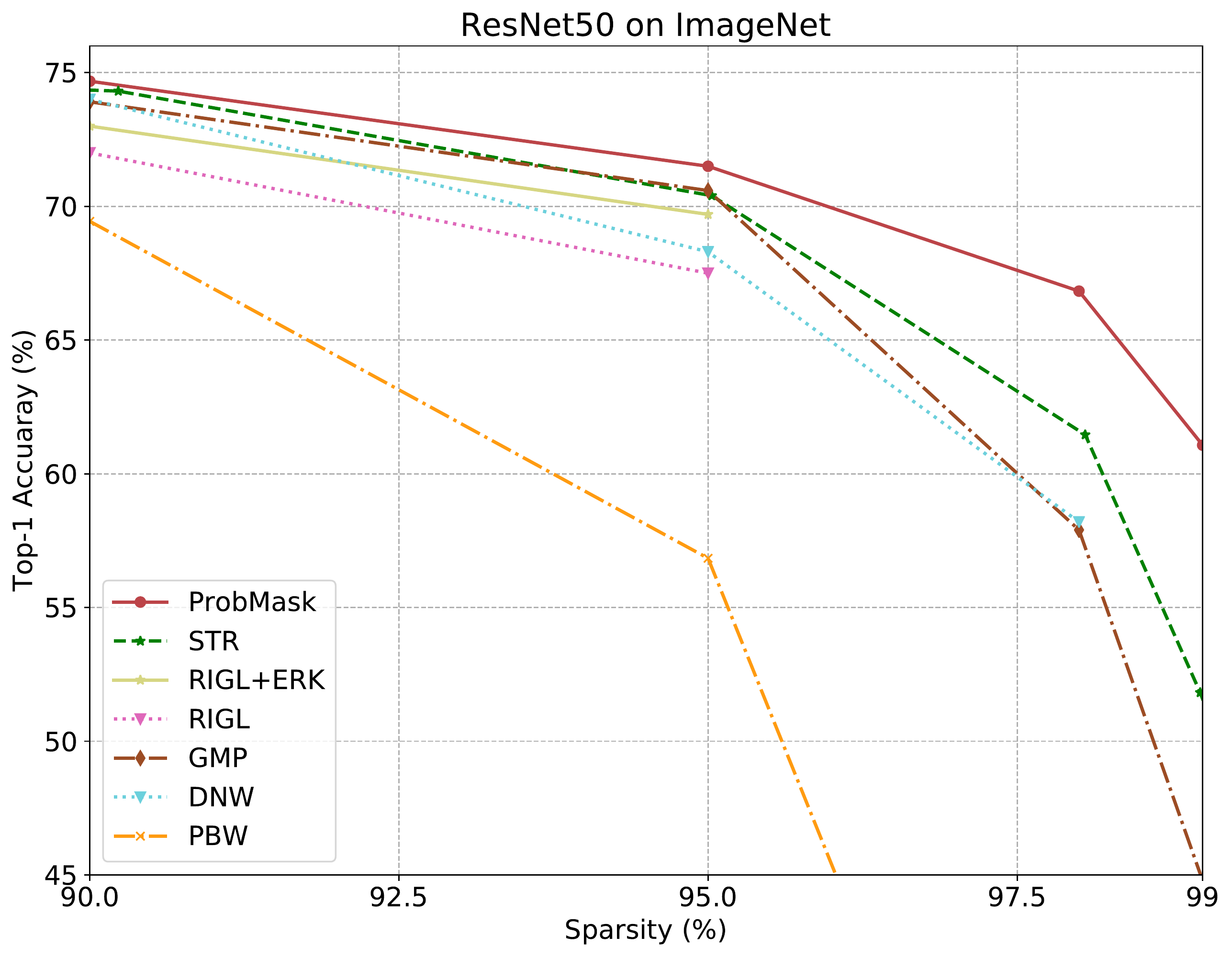}

\caption{\small{ProbMask comfortably beats state-of-the-art methods in all sparsity regions. Notably, the gap comes up to 5\% at 98\% sparsity and 10\% at 99\% sparsity.}}
\label{fig:results-imagenet}
\end{figure}

\begin{figure}[H]
\centering  
\includegraphics[scale=0.281]{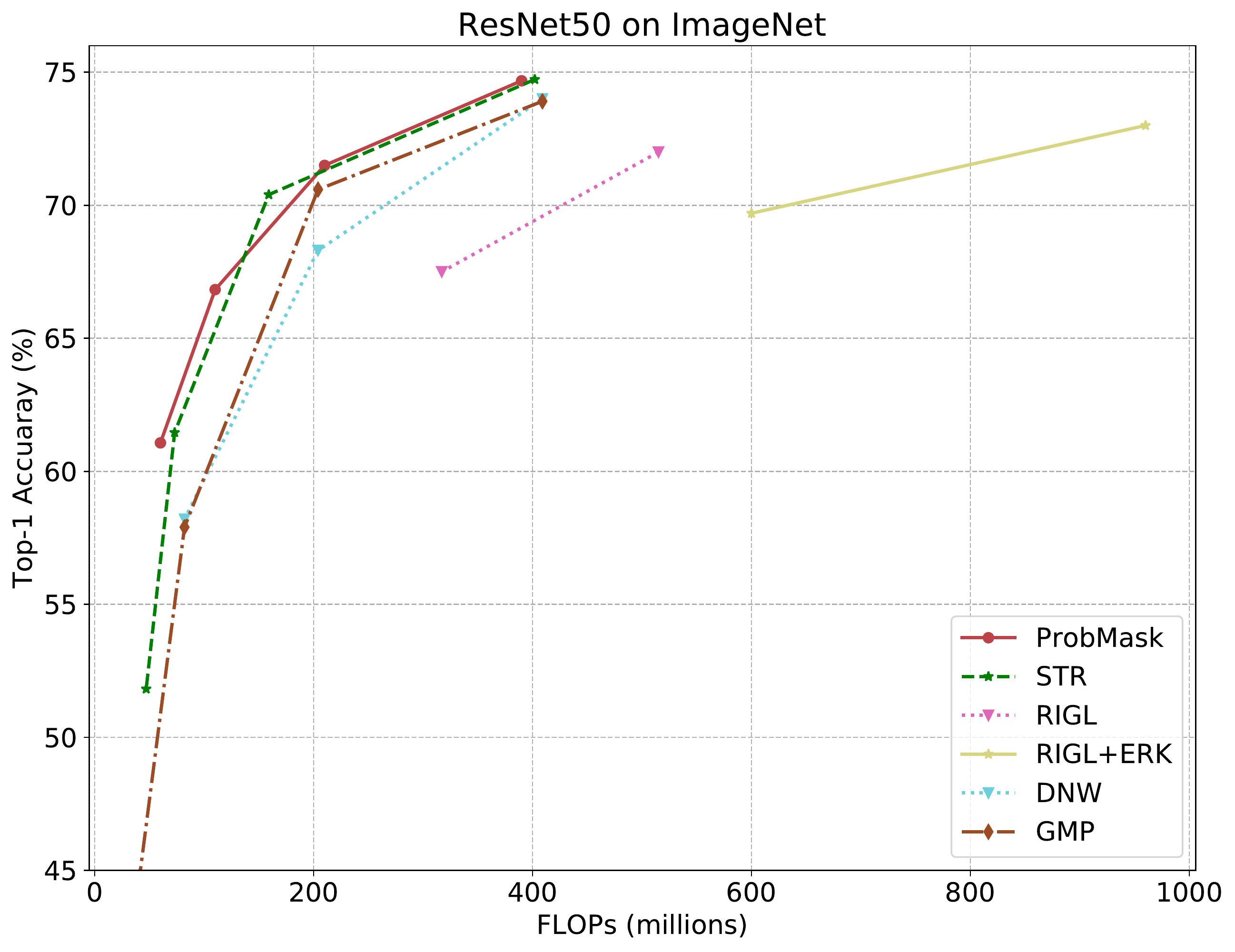}
% \vspace*{-8pt}
\caption{ProbMask obtains a smaller sparse network with comparable accuracy and FLOPs, still achieving state-of-the-art result on accuracy-vesus-FLOPs curve.}
\label{fig:results-imagenet_flop}
\end{figure}

In this section, we evaluate the performance of our ProbMask on ImageNet with ResNet50. 
%We choose PBW \cite{han2015deep}, MLPrune \cite{zeng2018mlprune}, GMP \cite{zhu2017prune}, STR \cite{kusupati2020soft}, RIGL \cite{wortsman2019discovering} as strong baselines. ProbMask is trained with 100 epochs following the practice of previous works. Basic parameters for weights like learning rate, momentum, weight decay follows the practice of STR \cite{kusupati2020soft}. Warmup, label smoothing(\cite{szegedy2016rethinking}) and regular data augmentation are applied as the same of STR \cite{kusupati2020soft}. We directly apply the same learning rate  with the experiments above for Adam to update the probability in ProbMask.
% For PBW and MLPrune, the pruned networks are finetuned with learning rate 1e-3 for 40 epochs.  %Open-source implementations, pre-trained models, and reported numbers of the available techniques are used as the baselines. 
% Although RIGL achieves better results with 5$\times$ running epochs, we just compare results around 100 epochs for time and resource limitation.%(要说吗？)
Table \ref{tab:result-imagenet} and Figure \ref{fig:results-imagenet} report the detailed accuracy at different pruning ratios. ProbMask steadily outperforms state-of-the-art methods with a large margin, especially when the pruning ratio is high than 98\%. Notably the gap comes up to 5\% at 98\% sparsity and 10\% at 99\% sparsity. DNW and GMP allocate uniform sparsity budget. They present reasonably good performance at 90\% sparsity while falling behind ProbMask by about 9\% at 98\% sparsity. This validates our previous claim that identifying weight allocation for different layers really matters. Uniform sparsity budget is a reasonable compromise but obviously don't give a perfect solution. STR attempts to learn weight allocation for different layers but don't give perfect results. ProbMask presents much better perfomance on high spasity regions, leading to a gap about 10\% percent at 99\% sparsity. With the global comparable nature of probability, ProbMask easily learns a much better weight allocation scheme for different layer. We also compare ProbMask with Sparse VD \cite{molchanov2017variational}  on sparsity 90\%. Sparse VD finds a subnet with 73.84\% Top-1 Acuucracy, a weaker result than ProbMask. We also observe noticable fluctuations between different runs, and this can be expected because Sparse VD adopts crude cut-off practice rather than sampling. This inevitably results in perfomance gap in training and testing phases. ProbMask learns a deterministic mask at the end of training, fixing the training and testing performance discrepancy problem. Figure \ref{fig:results-imagenet_flop} reports the accuracy-versus-FLOPs for ProbMask and compared methods. It shows that ProbMask finds a smaller mask with comparable accuracy and FLOPs and achieve state-of-the-art result on it.

\subsection{Powerful Tool for Finding Supermasks} \label{supermask}
\begin{figure}[htb!] 
\centering  
\includegraphics[scale=0.65]{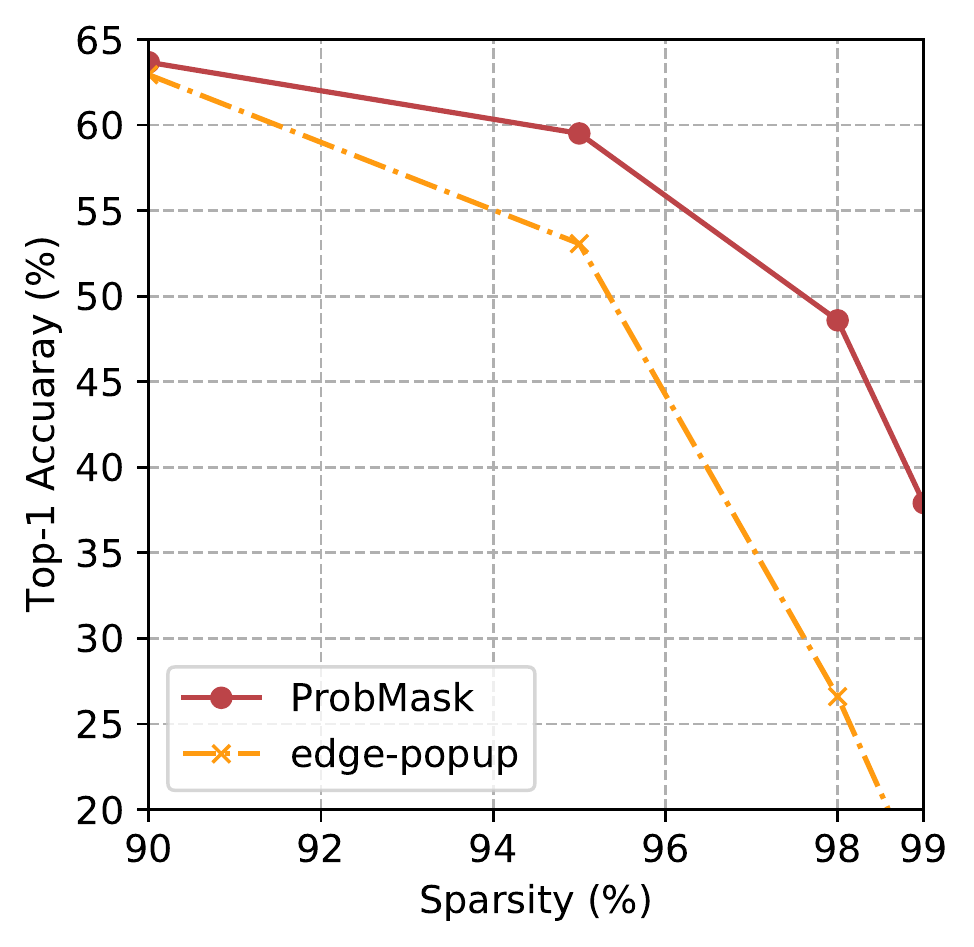}
\caption{ProbMask can find a supermask with just 2\% remaining weights and nearly 50\% accuracy on CIFAR-100. Weights are fixed at initialization state.}
\label{fig:supernet}
\end{figure}

Previous works on supermasks, i.e, subnetworks achieving good performance with weights fixed at random state, focus on sparsity region [10\%, 90\%]. Here, we would like to explore the performance of supermasks with higher sparsity, [90\%, 99\%]. We conduct experiments on modern architeture ResNet32 and dataset CIFAR-100, a harder task than CIFAR-10 where a large portion of previous experiments are conducted. In this experiment, weights are fixed at initialization state by Kaiming Normal \cite{he2015delving}. Hyperparameters follow the same as previous CIFAR experiments. According to Figure \ref{fig:supernet}, we observe that ProbMask easily scales to ultra sparse region with about 50\% accuracy and 2\% remaining weights, while state-of-the-art method edge-popup \cite{ramanujan2020s} collapse with less than 30\% accuracy. It is a surprising result that a subnet with 2\% fixed random weights still succeeds in obtaining nearly 50\% accuracy on a task with 100 categories. It shows that the structure in networks already provides  valuable information for classification.

\begin{table}[htb!]
\begin{center}
{\footnotesize
\begin{tabular}{p{2.5cm}<{\centering} p{0.8cm}<{\centering}p{0.8cm}<{\centering}p{0.8cm}<{\centering}p{0.8cm}<{\centering}}
\toprule
Dataset& & \multicolumn{2}{c}{ImageNet}& \\ \cmidrule(){1-5}
Ratio&  90\% & 95\% & 98\% & 99\%   \\ \cmidrule(r){1-1} \cmidrule(l){2-5}
ResNet50 & 77.01 & - & - & - \\  \cmidrule(r){1-1} \cmidrule(l){2-5}
 PBW\cite{han2015deep}&  69.44 & 56.84& 22.46 & 5.98 \\  \cmidrule(r){1-1} \cmidrule(l){2-5}
 MLPrune\cite{zeng2018mlprune} &  60.98 & 30.89 & 3.16 & 0.77 \\  \cmidrule(r){1-1} \cmidrule(l){2-5}
   GMP\cite{zhu2017prune} & 73.91 & 70.59 & 57.90 &44.78  \\  \cmidrule(r){1-1} \cmidrule(l){2-5}
 DNW\cite{wortsman2019discovering} &  74.00 & 68.30 & 58.20 & -\\  \cmidrule(r){1-1} \cmidrule(l){2-5}
 STR\cite{kusupati2020soft} &  74.31 & 70.40 & 61.46 & 50.35\\  \cmidrule(r){1-1} \cmidrule(l){2-5}
   RIGL\cite{menickrigging} &  72.00  & 67.50 & - & -  \\  \cmidrule(r){1-1} \cmidrule(l){2-5}
%   RIGL+ERK & 73.00 & 69.70 & - & - & -&-  \\  \cmidrule(r){1-1} \cmidrule(l){2-7}
%   RIGL & 72.00 & 67.50 & - & - & -&-  \\  \cmidrule(r){1-1} \cmidrule(l){2-7}
%   DNW & 74.00 & 68.30 & 58.2 & - & -&-  \\  \cmidrule(r){1-1} \cmidrule(l){2-7}
  ProbMask & \textbf{74.68} & \textbf{71.50} & \textbf{66.83} & \textbf{61.07} \\ 
 \bottomrule
\end{tabular}
}
\end{center}
\caption{Accuracy of ResNet50 on ImageNet at different pruning ratios. ProbMask steadily beats previous state-of-the-art methods on Hessian-based pruning, weight magnitude pruning, dense-to-sparse training and sparse-to-sparse training. RIGL improves with the help of ERK (Erdós-Rényi-Kernel) but will result in doubling the FLOPs at inference time, so we put it in Figure \ref{fig:results-imagenet}).}
\label{tab:result-imagenet}
\end{table}

% \begin{table}[H]

% \begin{center}
% {\footnotesize
% \begin{tabular}{p{1.6cm}<{\centering} p{0.7cm}<{\centering}p{0.7cm}<{\centering}p{0.7cm}<{\centering}p{0.8cm}<{\centering}}
% \toprule
% Dataset& & \multicolumn{2}{c}{CIFAR-100} &\\ \cmidrule(){1-5}
% Ratio& 90\% & 95\% & 98\% & 99\% \\ \cmidrule(r){1-1} \cmidrule(l){2-5} 
% ResNet32& \multicolumn{4}{c}{75.94(dense model)} \\  \cmidrule(r){1-1} \cmidrule(l){2-5}
%  edge-popup & 62.94 & 53.07 & 26.6 & 15.39  \\  \cmidrule(r){1-1} \cmidrule(l){2-5}
%  ProbMask &\textbf{63.67} &\textbf{59.51}& \textbf{48.59}& \textbf{37.91} \\
%  \bottomrule
% \end{tabular}
% }
% \end{center}
% \caption{}
% \end{table}

\section{Furthur Analysis}\label{sec:ablation-study}
% \vspace*{-10pt}
% In this section, we will show global comparability of probability adopted by ProbMask to measure weight importance, the superiority of our global sparsity constraint over layer-wise constraint, and the convergence of scores to an almost deterministic mask. 

\begin{figure}[htb!]
\centering  
\includegraphics[scale=0.42]{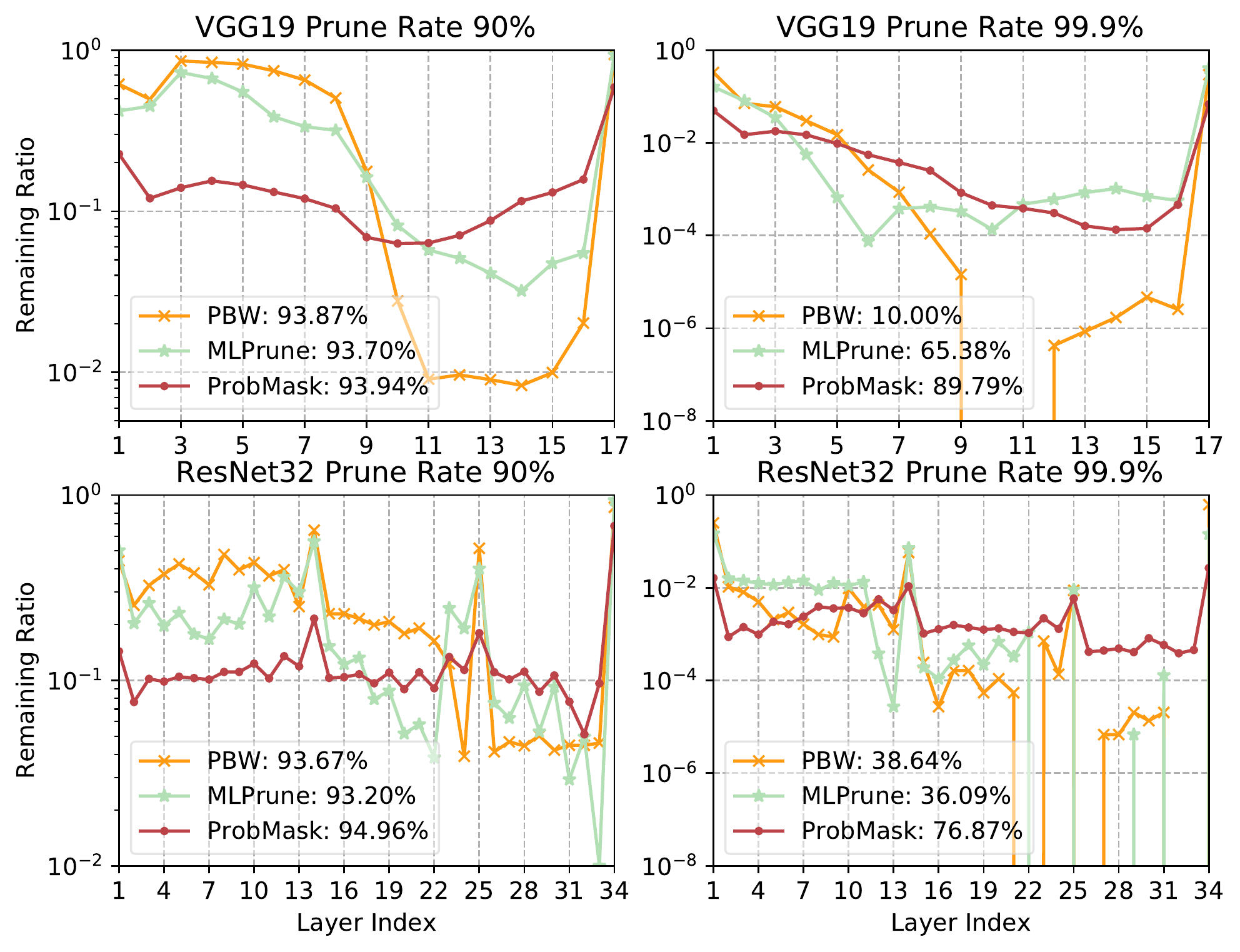}
% \vspace*{-5pt}
\caption{ProbMask learns a wise balance between uniform sparsity budget and abominably cutting one layer off, leading to compelling performance over sparsity range [90\%, 99.9\%].}
\label{fig:layer-remaining-ratio}
\end{figure}

\textbf{[Global Comparability of Probability]} Table \ref{tab:VGG19-CIFAR10-100} shows that PBW and MLPrune collapse  on CIFAR-10/100 when the pruning ratio is as high as 99.9\%. To explore the reason, we plot the remaining ratio across layers at pruning ratio of 90\% and 99.9\% on CIFAR-10  in Figure \ref{fig:layer-remaining-ratio}. It shows that when the pruning ratio is high, PBW and MLPrune prune almost all the weights in certain layers with remaining ratio approaching $10^{-6}$. The reason is that the proposed weight importance measure in PBW and MLPrune are not globally comparable. Although the weight importance scores in different layers have been normalized in MLPrune, their magnitudes are still quite different. A global threshold  could remove almost all  the weights in certain layers in order to achieve high enough pruning ratio. The pruning ratio of ProbMask varies in a proper range, attributed to the global comparable nature of probability. We observe that the learned sparsity budget is a wise balance between uniform sparsity budget and cutting one layer off. The first and last layer are assigned a bit more budget above average and several important bottleneck layers are detected automatically to assign more budget. 
% \vspace*{-10pt}
\begin{figure}[htb!]
\centering  
\includegraphics[scale=0.42]{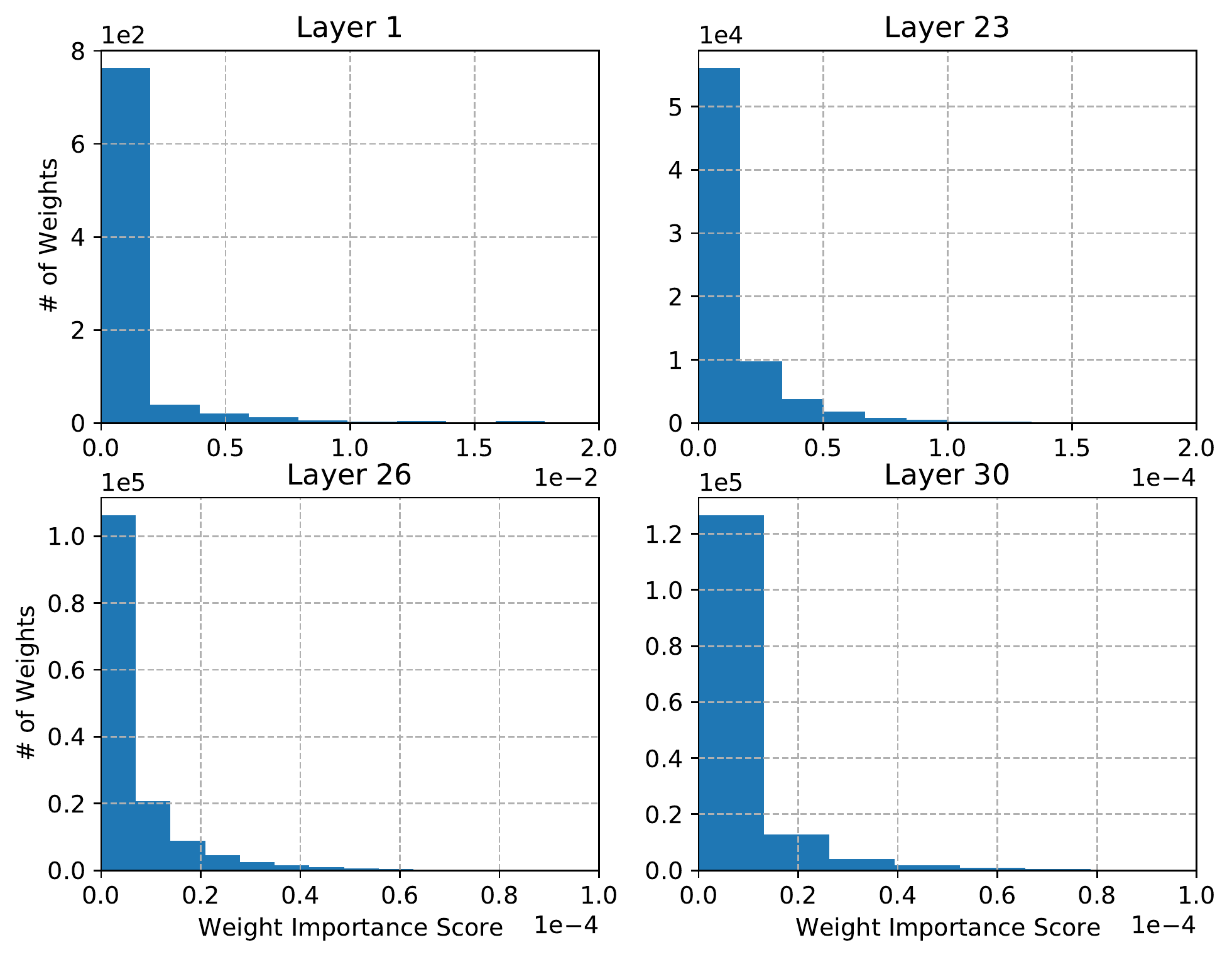}
\caption{Weight importance score histogram of ResNet32 from MLPrune with pruning rate 99.9\%. Note that the index of of x-axis is scaled to 1e-2 for Layer 1, and 1e-4 for Layer 23, 26, 30. This means that there exist two orders of magnitude difference across layers among weight importance scores.}
\label{fig:histogram_mlp}
\end{figure}

\textbf{[Superiority of Global Sparsity Constraint over Layer-wise Constraint]} In layer-wise constraint, we force all the pruning ratios in each layer to be equal and also equal to the one in the global constraint.  The experiment is conducted on CIFAR-10 with ResNet32 and the results are given in Table \ref{tab:superority-of-global-constraint}. It shows that the gap grows up rapidly when the pruning ratio is larger than 98\%. For example, when the pruning ratio is 99.9\%, the accuracy of global sparsity constraint can be up to 57.75\% higher than the layer-wise one. This points out the importance of identifying sparsity budget for different layers again by ablation study. %The non-uniform sparsity property attributes to the globally comparable nature of probabilities and play an important role pruning to extremely sparse regime. 
% \vspace*{-7pt}
% \begin{table}[htb!]
% \begin{center}
% {\footnotesize
% \begin{tabular}{p{1.2cm}<{\centering}p{0.7cm}<{\centering} p{0.7cm}<{\centering}p{0.7cm}<{\centering}p{0.7cm}<{\centering}p{0.7cm}<{\centering}p{0.7cm}<{\centering}}
% \toprule
% Dataset& & &  \multicolumn{2}{c}{CIFAR-10}&& \\ \cmidrule(){1-7}
% Ratio& 90\% &  95\% & 98\% & 99\% & 99.5\% & 99.9\%  \\ \cmidrule(r){1-1} \cmidrule(l){2-7}
% ResNet32&&\multicolumn{4}{c}{94.78 (dense model)}& \\  \cmidrule(r){1-1} \cmidrule(l){2-7}
%  LSB &  94.89 & 94.09 & 92.64 & 90.89 & 74.6 & 19.12 \\  \cmidrule(r){1-1} \cmidrule(l){2-7}
%  GSB &  \textbf{94.96} & \textbf{94.16} & \textbf{93.30} & \textbf{91.79} & \textbf{89.34} & \textbf{76.87}  \\  
%  \bottomrule

% \end{tabular}
% }
% \end{center}
% \caption{Comparing Layerwise Sparsity Budget (LSB) and Global Sparsity Budget(GSB) of ProbMask on ResNet32. GBS begins to take effect when sparsity comes up to 98\%.} \label{tab:superority-of-global-constraint}
% \end{table}

\begin{table}[htb!]
\begin{center}
{\footnotesize
\begin{tabular}{p{1.2cm}<{\centering}p{0.7cm}<{\centering} p{0.7cm}<{\centering}p{0.7cm}<{\centering}p{0.7cm}<{\centering}p{0.7cm}<{\centering}p{0.7cm}<{\centering}}
\toprule
Dataset& & &  \multicolumn{2}{c}{CIFAR-10}&& \\ \cmidrule(){1-7}
Ratio& 90\% &  95\% & 98\% & 99\% & 99.5\% & 99.9\%  \\ \cmidrule(r){1-1} \cmidrule(l){2-7}
ResNet32& 94.78 & - & - & - & - & -\\  \cmidrule(r){1-1} \cmidrule(l){2-7}
 LSB &  94.89 & 94.09 & 92.64 & 90.89 & 74.6 & 19.12 \\  \cmidrule(r){1-1} \cmidrule(l){2-7}
 GSB &  \textbf{94.96} & \textbf{94.16} & \textbf{93.30} & \textbf{91.79} & \textbf{89.34} & \textbf{76.87}  \\  
 \bottomrule

\end{tabular}
}
\end{center}
\caption{Comparing Layerwise Sparsity Budget (LSB, assigning uniform budget across layers) and Global Sparsity Budget (GSB) of ProbMask on ResNet32. GBS begins to take effect when sparsity comes up to 98\% and becomes prominent when sparsity is larger than 99.5\%.} \label{tab:superority-of-global-constraint}
\end{table}

\begin{figure}[htb!]
\begin{center}
\centering  
\includegraphics[scale=0.43]{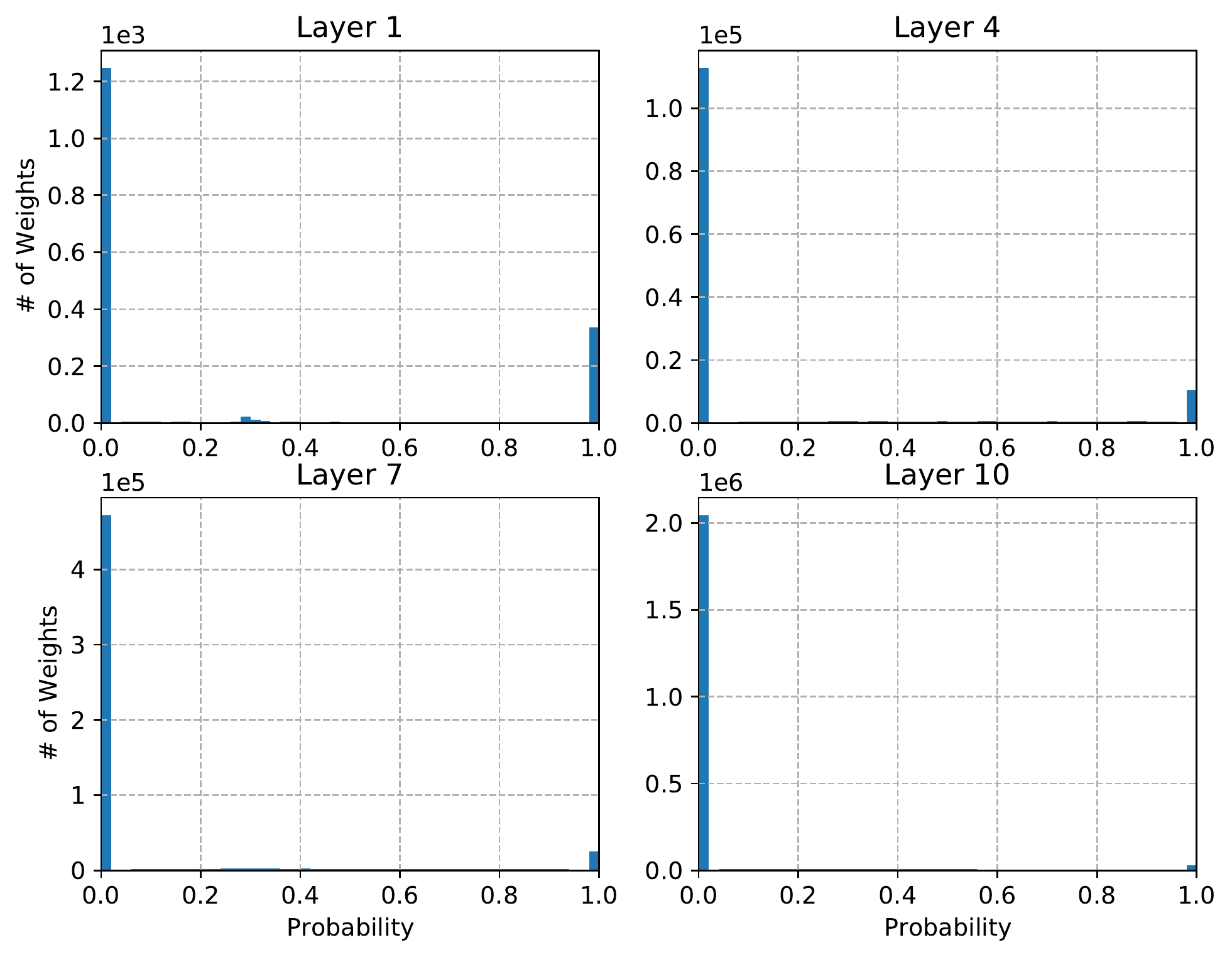}
\end{center}
    \caption{Probability histogram of VGG19 trained by ProbMask on CIFAR-10 at pruning rate 90\%.} 
\label{fig:probability-hist}
\end{figure}
% \vspace*{-10pt}
\textbf{[Convergence to Deterministic Mask]} To show that the mask trained by our ProbMask can converge to a deterministic mask after training, we randomly choose some layers from VGG19 and present their distribution of the probability value after training in Figure \ref{fig:probability-hist}. We can see that after training, almost all of the probabilities $\boldsymbol{s}_i$ can converge to either $0$ and $1$, leading to a deterministic  mask. This attributes to $\ell_1$ norm in our global sparsity constraint over the probability space and the precise chosen temperature annealing scheme.

\section{Conclusion}
This paper proposes an effective network sparsification method  ProbMask and demonstrates state-of-the-art results on various models and datasets. We provide evidence that probability can serve as a suitable global comparator to measure weight importance and solve the training and testing performance discrepancy problem observed in practice. ProbMask can also serve as a powerful tool for identifying subnetworks with high performance in a randomly weighted dense neural network.

% This formulation avoids the problem of tuning pruning rates individually for different layers manually or by using handcrafted heuristic rules, which was needed in previous approaches. Experimental results show ProbMask can outperform state-of-the-art with a large margin.

\section*{Acknowledgements}
    This work is supported by GRF 16201320.

%-------------------------------------------------------------------------

\newpage

{\small
\bibliographystyle{ieee_fullname}
\bibliography{egbib}
}
\clearpage
\newpage

\appendix

\twocolumn[
\begin{@twocolumnfalse}
	\begin{center}
	    \Large\bf
		\title{Effective Sparsification of Neural Networks with Global Sparsity Constraint	
		}
	\end{center}\end{@twocolumnfalse}
]

% \twocolumn[ \title{Appendix for \\ SparseBERT: Rethinking the Importance Analysis in Self-attention} ]

% \title{Effective Sparsification of Neural Networks with Global Sparsity Constraint}

% \maketitle
% \begin{center*}
% 	{\Large\bf Supplemental Material: How to Characterize The Landscape of\\ 
% 	\vspace{0.05in}
% 	Overparameterized Convolutional Neural Networks \par }
% \end{center*}

\section{Appendix}
In this appendix, we present additional MobileNetV1 \cite{howard2017mobilenets} experiment on ImageNet-1K, the general experimental configurations, proof for equation \ref{for proof}, proof for theorem \ref{theorem 1}, analysis on the effect of temperature annealing and PyTorch code snippets of ProbMask.
\vspace*{-3pt}
\subsection{MobileNetV1 on ImageNet-1K}
\begin{table}[htb!]
\begin{center}
{\footnotesize
\begin{tabular}{p{2.0cm}<{\centering}p{1.2cm}<{\centering} p{1.2cm}<{\centering}p{1.2cm}<{\centering}}
\toprule
Dataset& &ImageNet&\\ \cmidrule(){1-4}
Ratio& ProbMask & STR & GMP \\ \cmidrule(r){1-1} \cmidrule(l){2-4}
89\% & \textbf{65.19} & 62.10 & 61.80 \\  \cmidrule(r){1-1} \cmidrule(l){2-4}
94.1\% & \textbf{60.10} & 23.61 & - \\
 \bottomrule

\end{tabular}
}
\end{center}
\caption{ProbMask surpasses state-of-the-art methods by 3.09\% and 38.49\% Top-1 Accuracy, demonstrating the effectiveness and generalizability of ProbMask on lightweight MobileNetV1 \cite{howard2017mobilenets} architectures. Following the setting of \cite{kusupati2020soft}, 89\% and 94.1\% sparsity is chosen to compare at the same pruning rate.} \label{tab:superority-of-global-constraint-appendix}
\end{table}

\vspace*{-10pt}
\subsection{Experimental Configurations}\label{expsettings}
\begin{table}[H]
\begin{center}
{\footnotesize
\begin{tabular}{p{3.0cm}<{\centering} p{1.2cm}<{\centering}p{1.2cm}<{\centering}}
\toprule
Dataset & CIFAR & ImageNet\\ \cmidrule(){1-3}
GPUs & 1 & 4 \\ \cmidrule(){1-3}
Batch Size & 256 & 256 \\ \cmidrule(){1-3} 
Epochs & 300 & 100 \\ \cmidrule(){1-3} 
Weight Optimizer & SGD & SGD \\ \cmidrule(){1-3}
Weight Learning Rate & 0.1 & 0.256\\  \cmidrule(){1-3}
 Weight Momentum & 0.9 & 0.875  \\
\cmidrule(){1-3}
Probability Optimizer & Adam & Adam \\
\cmidrule(){1-3}
Probability Learning Rate & \textbf{6e-3} & \textbf{6e-3} \\
\cmidrule(){1-3}
$t_1$ & 48 & 16 \\
\cmidrule(){1-3}
$t_2$ & 180 & 60 \\
\cmidrule(){1-3}
Warmup & \xmark & \cmark \\
\cmidrule(){1-3}
Label Smoothing & \xmark & 0.1\\
\bottomrule

\end{tabular}
}
\end{center}
\caption{The bold-face probability learning rate 6e-3 is the \textbf{only} hyperparameter obtained by grid search on CIFAR-10 experiments on a small size network Conv-4 \cite{frankle2018lottery} and applied directly to larger datasets and networks. This demonstrates the generality of our proposed ProbMask to different datasets, different networks and different tasks, i.e., pruning networks and finding supermasks. Other hyperparameters are applied following the same practice of previous works \cite{ramanujan2020s, kusupati2020soft, liu2018rethinking, zhu2017prune}. The channels of ResNet32 for CIFAR experiments are doubled following the same practice of \cite{wang2020picking}. The temperature annealing scheme follows the same practice of \cite{xie2018snas}} 
\end{table}

\vspace{-5pt}
\subsection{Proof for equation \ref{for proof}}
\begin{proof}
The PDF (probability density function) of $\operatorname{Gumbel}(\mu, 1)$ is
\begin{equation}
f(z ; \mu)=e^{-(z-\mu)-e^{-(z-\mu)}}.
\end{equation}
The CDF (cumulative distribution function) of $\operatorname{Gumbel}(\mu, 1)$ is 
\begin{equation}
F(z ; \mu)=e^{-e^{-(z-\mu)}}.
\end{equation}
We just need to prove that 
\begin{equation}
\forall i, \, P\left(\log(s_i)-\log(1-s_i)+g_{1, i}-g_{2, i} \geq 0\right) = s_i.
\end{equation}
$g_{1,i}$ and $g_{2,i}$ are two $\operatorname{Gumbel}(0, 1)$ random variables sampled for $s_i$. The probability is taken with respect to $g_{1,i}$ and $g_{2,i}$. $s_i$ can be seen as a constant in the following proof.

Let $z_1 = \log(s_i)+g_{1, i}, z_2 = \log(1-s_i) + g_{2, i}.$ Then $z_1 \sim \operatorname{Gumbel}(\log(s_i), 1)$, $z_2 \sim \operatorname{Gumbel}(\log(1-s_i), 1)$.
\begin{align}
&P\left(\log(s_i)-\log(1-s_i)+g_{1, i}-g_{2, i} \geq 0\right) \\
=&P\left(z_2 \leq z_1\right) \\
=&\int_{-\infty}^{+\infty} \int_{-\infty}^{z_1} f(z_2; \log(1-s_i))f(z_1; \log(s_i)) dz_2dz_1 \\
% =&\int_{-\infty}^{+\infty} P(z_2 \leq z_1 | z_1)P(z_1) dz_1 \\
=&\int_{-\infty}^{+\infty} F(z_1; \log(1-s_i))f(z_1; \log(s_i)) dz_1 \\
=&\int_{-\infty}^{+\infty} e^{-e^{-(z_{1}-\log (1-s_i))}} \cdot e^{-(z_{1}-\log s_i)-e^{-(z_{1}-\log s_i)}}dz_{1}\\
=&\int_{-\infty}^{+\infty} e^{-e^{-z_{1}}(1-s_i)-z_{1}+\log s_i-e^{-z_{1}}s_i} d z_{1}\\
=&s_i \int_{-\infty}^{+\infty} e^{-e^{-z_{1}}-z_{1}} d z_{1} \\
=&s_i
\end{align}
$\int_{-\infty}^{+\infty} e^{-e^{-z_{1}}-z_{1}} d z_{1}$ is the integral of a Gumbel(0,1) random variable.
\end{proof}

\subsection{Proof for Theorem \ref{theorem 1}} 
\begin{proof}

The projection from $\bs{z}$ to set C can be formulated in the following optimization problem:
\begin{align}
    &\min_{\bs{s}\in \mathbb{R}^n} \frac{1}{2}\|\bs{s}-\bs{z}\|^2,\nonumber\\
    s.t. & \mathbf{1}\top \bs{s}  \leq K \mbox{ and } 0\leq \bs{s}_i \leq 1.\nonumber
\end{align}
Then we solve the problem with Lagrangian multiplier method.
\begin{align}
    L(\bs{s},v) &= \frac{1}{2}\|\bs{s}-\bs{z}\|^2 + v(\mathbf{1}^\top \bs{s} -K)\\
    &=\frac{1}{2}\|\bs{s}-(\bs{z}-v\mathbf{1})\|^2 + v (\mathbf{1}^\top \bs{z}-K) -\frac{n}{2}v^2.
\end{align}
with $v \geq 0 \mbox{ and } 0 \leq \bs{s}_i \leq 1$.
Minimize the problem with respect to $\bs{s}$, we have 
\begin{align}
    \tilde{\bs{s}} = \mathbf{1}_{\bs{z}-v\mathbf{1}\geq 1} + (\bs{z}-v\mathbf{1})_{1>\bs{z}-v\mathbf{1}>0}
\end{align}
Then we have
\begin{align}
g(v)=&L(\tilde{\bs{s}},v) \nonumber\\
   =& \frac{1}{2}\|[\bs{z}-v\mathbf{1}]_{-} + [\bs{z}-(v+1)\mathbf{1}]_{+}\|^2 \nonumber \\
   &+ v (\mathbf{1}^\top \bs{z}-s) -\frac{n}{2}v^2 \nonumber  \\
=&\frac{1}{2}\|[\bs{z}-v\mathbf{1}]_{-}\|^2 +\frac{1}{2}\|[\bs{z}-(v+1)\mathbf{1}]_{+}\|^2\nonumber \\
&+ v (\mathbf{1}^\top \bs{z}-s) -\frac{n}{2}v^2,  v\geq 0. \nonumber \\
g'(v)=& \mathbf{1}^\top [v\mathbf{1}-\bs{z}]_{+} +\mathbf{1}^{\top} [(v+1)\mathbf{1}-\bs{z}]_{-}\nonumber \\
&+(1^T\bs{z}-s)-nv \nonumber \\
    % =& \mathbf{1}^\top[(\bs{z}-v\mathbf{1}).\mathbf{clamp}(0,1)] - K
    % ,v\geq 0.\\
=&\mathbf{1}^\top\min (1, \max(0, \bs{z}-v\mathbf{1})) - K,v\geq 0.\nonumber
\end{align}
It is easy to verify that $g'(v)$ is a monotone decreasing function with respect to $v$ and we can use a bisection method solve the equation $g'(v) = 0$ with solution $v^*_1$. Then we get that $g(v)$ increases in the range of $(-\infty, v^*_1$] and decreases in the range of $[v^*_1, +\infty)$. The maximum of g(v) is achieved at 0 if $v^*_1 \leq 0$ and $v^*_1$ if $v^*_1 >0$. Then we set $v^*_2 = max(0, v^*_1)$. Finally we have 
\begin{align}
    \bs{s}^* =& \mathbf{1}_{\bs{z}-v_2^*\mathbf{1}\geq 1} + (\bs{z}-v_{2}^*\mathbf{1})_{1>\bs{z}-v_2^*\mathbf{1}>0}\\ 
    =&\min (1, \max(0, \bs{z}-v_{2}^{*}\mathbf{1})).
\end{align}
\end{proof}

\subsection{Temperature Annealing}
Thanks to the $\ell_1$ norm and cube $[0,1]^n$ in our constraint, most probabilities will converge to 0 or 1 at the end of training, which is shown in 
\begin{align}
\bs{s} = \min (1, \max(0, \bs{z}-v_{2}^{*}\mathbf{1})). \nonumber
\end{align}
Traditional temperature annealing starts with a relative high value, i.e., 1 to have a smooth relaxation and gradually decrease to a small value to make relaxation close to the original objective function. In this section we analyze how the temperature annealing contributes to the training process, especially helping probabilities converge to 0 or 1.

Firstly consider the gradient:
\begin{align}
&\nabla_{s_i} \mathcal{L}\left(\boldsymbol{w}, \sigma\Big(\frac{\log(\frac{\boldsymbol{s}}{\boldsymbol{1}-\boldsymbol{s}})+\boldsymbol{g_1}-\boldsymbol{g_0}}{\tau}\Big)\right) \\
%= &\nabla_{\sigma_i} \mathcal{L}\left(\boldsymbol{w}, \sigma\left(\frac{\log(\frac{\boldsymbol{s}}{\boldsymbol{1}-\boldsymbol{s}})+\boldsymbol{g_1}-\boldsymbol{g_0}}{\tau}\right)\right)\nabla_{s_i}\sigma\left(\frac{\log(\frac{s_i}{1-s_i})+g_{1,i}-g_{0,i}}{\tau}\right) \label{term T}.
% = &\nabla_{\sigma_i} \mathcal{L}\left(\boldsymbol{w}, \sigma\Big(\frac{\log(\frac{\boldsymbol{s}}{\boldsymbol{1}-\boldsymbol{s}})+\boldsymbol{g_1}-\boldsymbol{g_0}}{\tau}\Big)\right)\nabla_{s_i}\sigma\left(u\left(s_i\right)\right) \label{term T}, \\
= &\nabla_{\sigma_i} \mathcal{L}\left(\boldsymbol{w}, \sigma\Big(\frac{\log(\frac{\boldsymbol{s}}{\boldsymbol{1}-\boldsymbol{s}})+\boldsymbol{g_1}-\boldsymbol{g_0}}{\tau}\Big)\right)S\label{term T},
\end{align}
where $S = \nabla_{s_i}\sigma(\frac{\log(\frac{s_i}{1-s_i})+g_{1,i}-g_{0,i}}{\tau})$. We can see that the larger the magnitude $|S|$, $z_i$ (step 8 in Algorithm \ref{alg:SST}) will vary more greatly.

Take $x = \frac{1}{\tau} \in [1, +\infty)$, $ r= log(s_i)-log(1-s_i)+g_{1,i}-g_{0,i}$. We have
\begin{align}
% &\nabla_{s_i}\sigma\left(\frac{\log(s_i)-\log(1-s_i)+g_{1,i}-g_{0,i}}{\tau}\right)\\
S = \frac{\sigma(rx)(1-\sigma(rx))x}{s_i(1-s_i)}
% = &\nabla_{s_i}\sigma(rx) \\
\end{align}
Since $S$ is an even function w.r.t $r$, we just consider the case $r > 0$. Then we take the gradient w.r.t to $x$.
\begin{align}
&\nabla_{x}\left(\sigma(rx)(1-\sigma(rx))x\right)\\
= &\sigma(rx)(1-\sigma(rx))(rx-2rx\sigma(rx)+1)
\end{align}
Take $y = rx, y > 0$ since $x>0$ and $r>0$. The solution to $y-2y\sigma(y)+1 = 0$ is around 1.55. S is a monotonically increasing function for $x \in (0, \frac{1.55}{r}]$ and monotonically decreasing function for $x \in [\frac{1.55}{r}, +\infty)$.

\begin{figure}[htb!] 
\centering  
\includegraphics[scale=0.45]{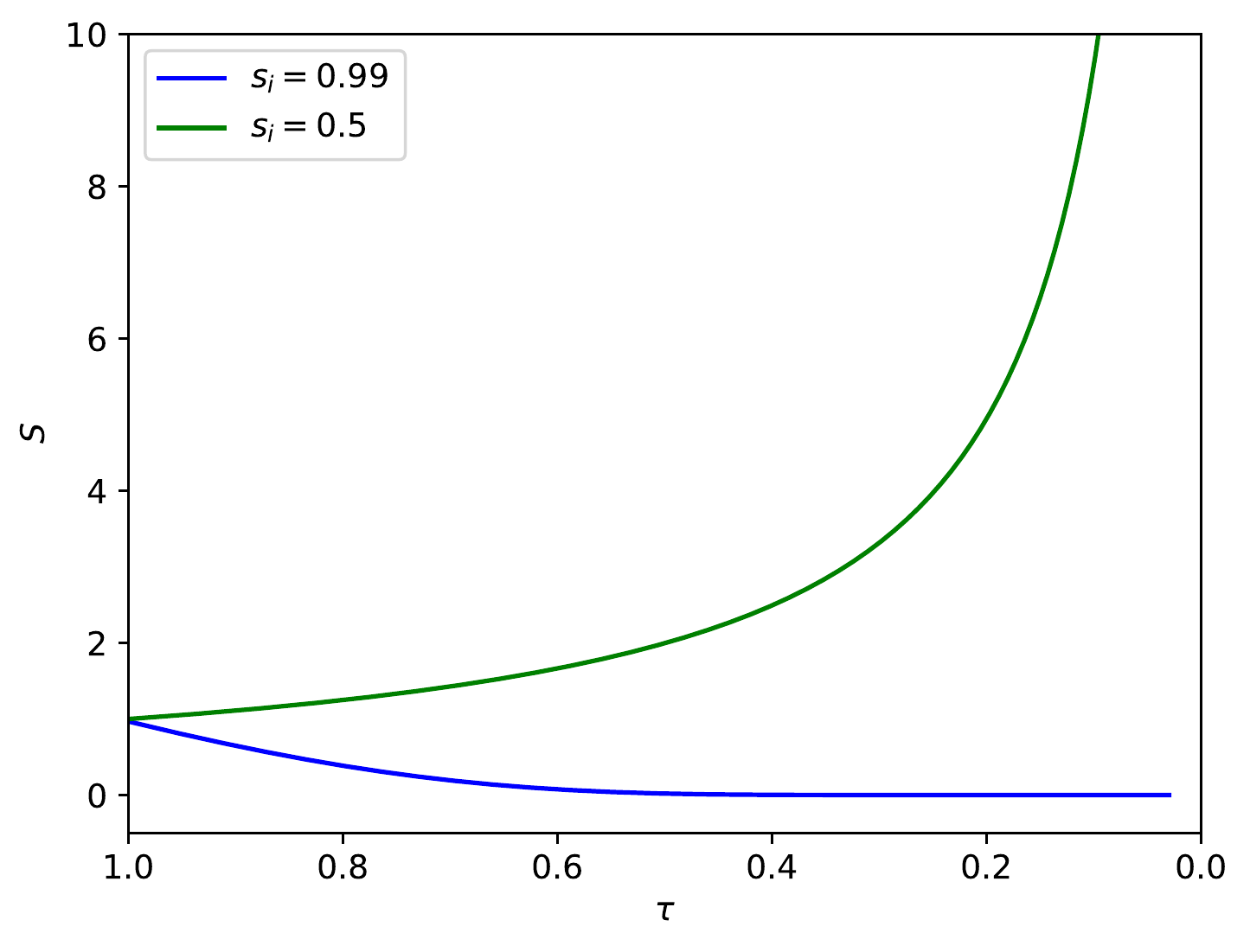}
\caption{The value of S changes with $\tau$ approching zero.}
\label{fig:func_t}
\end{figure}

Now we analyze S using two special cases. Take $g = g_{1,i}-g_{2,i}$, $g \sim Logistic(0, 1)$. Take $s_i = 0.99 \mbox{ and } g = 0.04$ for example. $r = \log(99) + 0.1 \approx 4.63$. S is monotonically decreasing function for $x \in [1, +\infty)$. Take $s_i = 0.5 \mbox{ and } g = 0.04$ for example. S would increase as $\tau$ decreases to 0.03, since we set $\tau = 0.97(1-t/T)+0.03$. We plot the corresponding graph in Figure \ref{fig:func_t}.

From the above two examples, we know that for probabilities around 0.5, S becomes larger in the training process, potentially making $|z_i|$ large and finally make probability come close to 0 or 1 after projection. For probabilities close to 0 or 1, S becomes smaller in the training process, making them stay close to 0 or 1 at the end of training.

\begin{lstlisting}[float=*t, language=python, caption={PyTorch Code Snippets for ProbMaskConv.}]
class ProbMaskConv(nn.Conv2d): 
    def __init__(self, *args, **kwargs):
        super().__init__(*args, **kwargs)
        self.scores = nn.Parameter(torch.Tensor(self.weight.size()))    # Probability
        self.subnet = None                                              # Mask
        self.scores.data = (torch.ones_like(self.scores)* parser_args.score_init_constant)

    def forward(self, x):               # Sample a mask and forward propagation
        if not parser_args.discrete:    # Training
            eps = 1e-20
            temp = parser_args.T
            uniform0 = torch.rand_like(self.scores)
            uniform1 = torch.rand_like(self.scores)
            noise = -torch.log(torch.log(uniform0 + eps) / torch.log(uniform1 + eps) + eps)
            self.subnet = torch.sigmoid((torch.log(self.scores + eps) - torch.log(1.0 - self.scores + eps) + noise) * temp)
        else:                           # Testing
            self.subnet = (torch.rand_like(self.scores) < self.scores).float()
        w = self.weight * self.subnet
        x = F.conv2d(x, w, self.bias, self.stride, self.padding, self.dilation, self.groups)
        return x



\end{lstlisting}
\begin{lstlisting}[float=*t, language=python, caption={PyTorch Code Snippets for Projection in Theorem \ref{theorem 1}}]
def constrainScoreByWhole(model):  
    total = 0
    for n, m in model.named_modules():
        if hasattr(m, "scores"):
            total += m.scores.nelement()
    v = solveV(model, total)                # Calculate v_2^* in Theorem 1
    for n, m in model.named_modules():
        if hasattr(m, "scores"):
            m.scores.sub_(v).clamp_(0, 1)   # Do the projection 

def solveV(model, total):                   # Solve solution to Equation 7 with bisection search
    k = total * parser_args.prune_rate
    a, b = 0, 0
    for n, m in model.named_modules():
        if hasattr(m, "scores"):
            b = max(b, m.scores.max())
    def f(v):
        s = 0
        for n, m in model.named_modules():
            if hasattr(m, "scores"):
                s += (m.scores - v).clamp(0, 1).sum()
        return s - k
    if f(0) < 0:
        return 0
    itr = 0
    while (1):
        itr += 1
        v = (a + b) / 2
        obj = f(v)
        if abs(obj) < 1e-3 or itr > 20:
            break
        if obj < 0:
            b = v
        else:
            a = v
    v = max(0, v)
    return v
\end{lstlisting}

\end{document}